\documentclass[twoside,11pt]{article}

\usepackage{jmlr2e}

\usepackage{url}            \usepackage{booktabs}       \usepackage{amsfonts}       \usepackage{nicefrac}       
\usepackage{mathtools}
\usepackage{amssymb}

\usepackage{graphicx}
\usepackage{grffile}
\usepackage{subcaption}
\usepackage[percent]{overpic}
\usepackage{multirow}

\usepackage{times}
\usepackage[varqu,varl,var0,scaled=0.97]{inconsolata}
\usepackage[protrusion=true,expansion=true]{microtype}

\usepackage[dvipsnames]{xcolor}
\usepackage{cleveref}
\usepackage{hyperref}
\newcommand\myshade{90}
\colorlet{mylinkcolor}{NavyBlue}
\colorlet{mycitecolor}{Aquamarine}
\colorlet{myurlcolor}{Aquamarine}
\hypersetup{
  linkcolor  = mylinkcolor!\myshade!black,
  citecolor  = mycitecolor!\myshade!black,
  urlcolor   = myurlcolor!\myshade!black,
  colorlinks = true,
}

\newcommand{\alllayerstitle}{Are All Layers Created Equal?}

\newcommand{\fcn}[1]{\textrm{FCN}{#1}}

\newcommand{\vgg}[1]{\textrm{VGG}{#1}}
\newcommand{\resnet}[1]{\textrm{ResNet}{#1}}

\usepackage{xspace}
\newcommand{\mnist}{MNIST\xspace}
\newcommand{\cifar}{CIFAR-10\xspace}
\newcommand{\imagenet}{ImageNet\xspace}

\newcommand{\robust}{robust\xspace}
\newcommand{\critical}{critical\xspace}

\firstpageno{1}
\usepackage{lastpage}
\jmlrheading{23}{2022}{1-\pageref{LastPage}}{1/20; Revised
2/22}{2/22}{20-069}{Chiyuan Zhang, Samy Bengio, Yoram Singer}
\ShortHeadings{\alllayerstitle}{Zhang, Bengio and Singer}

\begin{document}

\title{\alllayerstitle}

\author{\name Chiyuan Zhang \email chiyuan.zh@gmail.com        \AND
  \name Samy Bengio \email bengio@gmail.com        \AND
  \name Yoram Singer \email yoram.singer@gmail.com      }

\editor{Lorenzo Rosasco}

\maketitle

\begin{abstract}	Understanding deep neural networks is a major research objective with
	notable experimental and theoretical attention in recent years.  The
	practical success of excessively large networks underscores the need for
	better theoretical analyses and justifications. In this paper we focus on
	layer-wise functional structure and behavior in overparameterized deep
	models. To do so, we study empirically the layers' robustness to
	post-training \emph{re-initialization} and \emph{re-randomization} of the
	parameters.  We provide experimental results which give evidence for the
	heterogeneity of layers.  Morally, layers of large deep neural networks can be
	categorized  as either ``\robust'' or ``\critical''.  Resetting the
	\robust layers to their initial values does {\em not} result in adverse
	decline in performance. In many cases, \robust layers hardly change
	throughout training. In contrast, re-initializing \critical layers
	vastly degrades the performance of the network with test error essentially
	dropping to random guesses. Our study provides further evidence that mere
	parameter counting or norm calculations are too coarse in studying
	generalization of deep models, and ``flatness'' and robustness analysis of
	trained models need to be examined while taking into account the
	respective network architectures.
\end{abstract}


\section{Introduction}

Deep neural networks have been remarkably successful in many real world
machine learning applications. The practical success of excessively large
networks cannot be explained by the classical wisdom of uniform convergence
and learnability.  In many critical applications, such as self-driving
vehicles and automatic medical diagnostics, distilled understanding of the
systems can be as important as achieving the state-of-the-art performance.
One important question is on interpreting and explaining the decision
function of trained networks.  It is closely related to another important
topic on networks' generalization and robustness under drifting or even
adversarially perturbed data distribution. In this paper, we study how
individual layers coordinate the computation in trained neural network
models, and relate the empirical results to generalization and robustness
properties.

Theoretical research of the functions computed by neural networks dates back
to the '80s. It is known that a neural network with a single (sufficiently
wide) hidden layer is a universal approximator for continuous functions over
compact domains~\citep{cybenko1989approximation, hornik1989multilayer,
hornik1991approximation, devroye2013probabilistic, gyorfi2006distribution,
anthony2009neural}.  More recent research further examines whether
\emph{deep} networks can have superior representation power than
\emph{shallow} ones with the same number of units or
edges~\citep{Pinkus:1999gk,  Delalleau:2011vh, montufar2014number,
pmlr-v49-telgarsky16, Shaham:2015uf, Eldan:2015uc,
DBLP:journals/corr/MhaskarP16, rolnick2017power}. The capacity to represent
arbitrary functions on finite samples is also extensively discussed
\citep{hardt2016identity, zhang2016understanding, nguyen2018optimization,
yun2018finite}. However, the constructions used in the aforementioned work
for building networks approximating particular functions are typically
``artificial'' and are unlikely to be obtained by gradient-based learning
algorithms.  We focus instead on empirically studying, \emph{post}-training,
the role different layers take in representing a learned function by
gradient-based methods.

Generalization is a fundamental theoretical question in machine learning.
The recent observation that big neural networks can fit random labels on the
training set \citep{zhang2016understanding} makes it difficult to apply
classical learning theoretic results based on uniform convergence over the
hypothesis space. One approach to get around this issue is to show that,
while the space of neural networks of a given architecture is huge,
gradient-based learning on ``well behaved'' tasks leads to relatively
``simple'' models. More recent research focuses on the analysis of the
post-training complexity metrics such as \emph{norm}, \emph{margin},
\emph{robustness}, or \emph{flatness} of the learned model in contrast to
the pre-training \emph{capacity} of the entire hypothesis space. This line
of work obtained sharper generalization bounds for for neural networks
\citep[e.g.][]{kawaguchi2017generalization, bartlett2017spectrally,
neyshabur2017exploring, liang2017fisher}.  In another line of work, Belkin
et al. show that overparameterization and perfectly fitting the training set
(i.e. \emph{interpolation}) is not an issue for generalization
\citep[e.g.][]{belkin2018reconciling, belkin2018overfitting,
belkin2018understand, azizan2019stochastic}.  Further more, they argue that
even norm-based generalization bounds could become non-informative in the
regime of interpolation.  Our work provides further empirical evidence and
alludes to more fine-grained analysis.  We propose that the layers in a deep
network are not homogeneous in the role they play at representing a
predictor. Some layers are \critical to forming good predictions while
others are \robust as they are fairly insensitive to the assignment of their
weights during training.  Thus, depending on the capacity of the network and
the complexity of the target function, gradient-based trained networks
conserve the complexity by not using excess capacity.

Before proceeding, we would like to further mention a few related papers.
Modern neural networks are typically overparameterized and thus redundant in
their representations. Previous work exploited overparameterization to
compress \citep{han2015deep} or distill \citep{hinton2015distilling} a
trained network. It is also shown that one can achieve comparable
performance by training only a small fraction of network parameters such as
a subset of the channels in each convolutional
layer~\citep{Rosenfeld:2018wf}. As a tool for interpreting residual networks
as ensemble of shallow networks, \citet{veit2016residual} found that
residual blocks in a trained network can be deleted or permuted to some
extent without degrading the test performance too much. Another line of
research showed that under extreme overparameterization, such as when the
network width is polynomial in the training set size and input dimension
\citep{Allen-Zhu2018-eq, Du2018-xp, Du2018-kc, Zou2018-lp}, or even in the
asymptotic regime of infinite width \citep{jacot2018neural,lechao2019}, the
network weights move slowly during training. We make similar observations in
this paper. However, we find that in more pragmatic settings (network widths
in the order of thousands), different layers exhibit different behaviors and
the network cannot be treated in a monolithic way.

Notice that our work, as well as the literature cited above, focuses on
conventional neural networks, where the output of the optimization process
is a single set of parameters computing a deterministic function. There is
an important, though less related, line of work that considers stochastic
neural networks where the output of the training process is a
\emph{distribution over model parameters}. Under this setting,
generalization bounds are derived with PAC-Bayes
analysis~\citep[e.g.][]{Neyshabur2018-zn, Arora2018-ou, Zhou2019-ii,
rivasplata2019pac}, and in some cases bounds with non-vacuous values can be
computed numerically \citep[e.g.][]{Dziugaite2017-bi, rivasplata2019pac}. As
will be shown in Section~\ref{sec:other-robustness}, the layer robustness
properties identified in this paper can be used to turn a deterministic
model trained with the conventional methods into a stochastic model.

The rest of the paper is organized as follows. Our experimental framework
and notions of robustness to modifications of layers are introduced in
Section~\ref{sec:setting}. Section~\ref{sec:layer-wise-analysis} presents the
results and analysis of layer robustness for a wide range of neural
network models. Section~\ref{sec:joint-robustness} presents experiments with 
joint robustness. Section~\ref{sec:other-domains} shows the generalizability
of the phenomenon in alternative domains and architectures. In Section~\ref{sec:other-robustness}
we discuss connections to other notions of robustness. Finally, the paper ends on
Section~\ref{sec:discussion} with a discussion and summarization of our main contributions.

\section{Setting}
\label{sec:setting}

We focus on feed forward networks which consist of multiple \emph{layers}
where each unit in a layer takes inputs from units in the previous layer.
Let $\mathcal{F}^D=\{f_{\theta}: \theta \in \Theta_1\times\cdots\times\Theta_D\}$
be the function space of a (particular) neural network architecture with $D$
(parametric) layers. Each admissible $\theta$ is a list
$\theta=(\theta_1,\ldots,\theta_D)$ with $\theta_d$ from $\Theta_d$ for all
$d \in [D]$. We are interested in analyzing \emph{post-training}
characteristics of layers used in popular deep networks. Such networks are
typically trained using stochastic gradient descent (SGD) which initialize
the parameters $\theta^0 = (\theta_1^0, \ldots, \theta_D^0)$ by sampling
from a pre-defined distribution $\mathcal{P}_d$ over $\Theta_d$. The choice
of $\mathcal{P}_d$ typically depends on structural properties such as fan-in, 
and fan-out of each layer. In our experiments, we take the default 
initialization schemes used in open source deep learning libraries.
After training for $T$ epochs, the parameters of the last epoch
$\theta^{T} = (\theta_1^{T}, \ldots, \theta_D^{T})$ are used as the final
trained model.

In a classification neural network, the decision function $f_{\theta^T}$ maps
an input to a class from a finite set of labels. Performance of the trained
networks is measured in terms of the agreement between its predicted labels and
the true labels on a newly observed test set.  Unless noted otherwise, we
use the term \emph{performance} to designate the 0-1 classification accuracy.
Our study of the layer structure evaluates the performance along the trajectory
$\theta^{\tau}$ throughout the entire sequence of training epochs $\tau \in [T]$.

\paragraph{Checkpointing.}
We save the model parameters at the end of every epoch
$\tau\in[T]$ during training and call the saved models \emph{checkpoints}.
Checkpoint-$T$ consists of the parameters for the final model. There is also a
special checkpoint-$0$ containing random weights initialized \emph{before}
the training starts.

A deep network constructs a \emph{representation} of its inputs by
incrementally applying transformations defined by each layer. Each layer
consists of a linear transformation on its inputs, matrix multiplication,
followed by nonlinear activation functions applied to the result of the
matrix multiplication. Notable examples are the sigmoid and the Rectified Linear
Unit (ReLU) activations. As a result, the representation at a particular layer
recursively depends on all the layers beneath. This complex dependency
makes it challenging to isolate and inspect each layer independently in
theoretical studies. In this paper, we introduce and use the following two
empirical probes to inspect the individual layers in a trained neural network.

\paragraph{Re-initialization.} After training concludes, for each layer
$d\in[D]$ separately, we \emph{re-initialize} its parameters by assigning
$\theta_d^{T}\leftarrow \theta_d^{0}$ while keeping the rest of the
parameters intact. We thus obtain $D$ different models where each model is
of the form $(\theta_1^{T},\ldots, \theta_{d-1}^{T}, \theta_d^{0},
\theta_{d+1}^{T}, \ldots, \theta_D^{T})$. The models are then evaluated on
the test set.
The relation of a layer to the effect on performance of re-initializing
it is referred to as the \emph{re-initialization robustness} of the layer.
Here $\theta_d^{0}$ denotes the randomly initialized values loaded from checkpoint-$0$.
More generally, for each epoch $\tau \in [T]$, we can
\emph{re-initialize} the $d$'th layer to its value on epoch $\tau$ through
the assignment $\theta_d^{T}\leftarrow \theta_d^{\tau}$, and study the
\emph{re-initialization robustness} of layer $d$ w.r.t checkpoint-$\tau$.

\paragraph{Re-randomization.} As one step further, \emph{re-randomization}
of a layer $d$ refers to assigning random parameters $\tilde{\theta}_d$ by
re-sampling using the same distribution $\mathcal{P}_d$ used for the
initialization of $\theta_d^0$, namely $\theta_d^{T}\leftarrow\tilde{\theta}_d$,
and evaluating $(\theta_1^{T},\ldots,\theta_{d-1}^{T},\tilde{\theta}_d,\theta_{d+1}^{T},\ldots,
\theta_D^{T})$.
Analogously, \emph{re-randomization robustness} of a layer $d$ is the
relation of that layer to the effect on performance of re-randomizing it.

We emphasize that \emph{no} re-training or fine-tuning is
performed after a network is re-initialized/re-randomized.  When a network
exhibits negligible\footnote{We do not quantify how much ``negligible'' is,
as we believe there is no universal threshold across all models and tasks.
Our empirical results indicates that there is no or little ambiguity in
categorizing layers due to the sharp difference in performance of \robust
and \critical layers.} decrease in performance after re-initializing or
re-randomizing of a layer, we say that the layer is \emph{\robust} and
otherwise the layer is called \emph{\critical}.

\section{Robustness of Individual Layers} \label{sec:layer-wise-analysis}
The datasets we use in our robustness study are standard image
classification benchmarks: \mnist, \cifar{}, and \imagenet. All networks
were trained using SGD with momentum using a piecewise constant learning rate
schedule. See Appendix~\ref{app:exp-details} for further details.

\subsection{Fully Connected Networks} \label{sec:mnist-mlp}
\begin{figure*}
  \begin{subfigure}{.32\linewidth}
    \includegraphics[width=\linewidth]{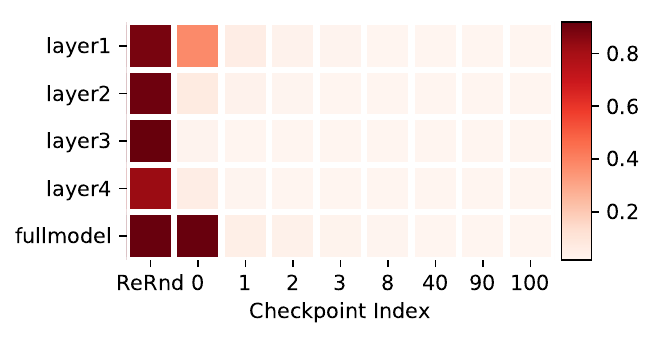}
    \caption{Test error}
  \end{subfigure}
  \begin{subfigure}{.32\linewidth}
    \includegraphics[width=\linewidth]{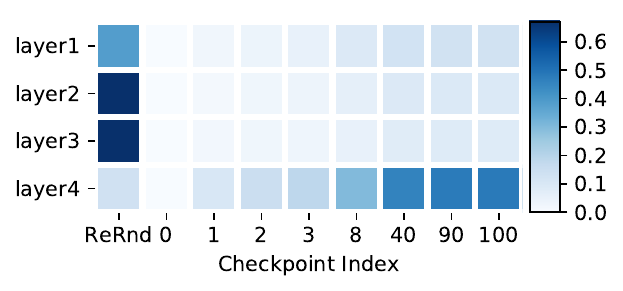}
    \caption{$\|\theta_d^{\tau}-\theta_d^{0}\|$}
  \end{subfigure}
  \begin{subfigure}{.32\linewidth}
    \includegraphics[width=\linewidth]{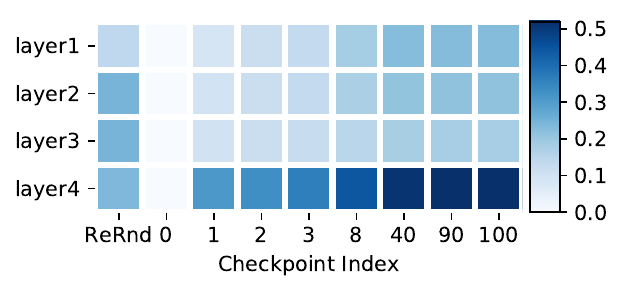}
    \caption{$\|\theta_d^{\tau}-\theta_d^{0}\|_\infty$}
  \end{subfigure}
  \caption{
    \textbf{Robustness results for \fcn{~$3\times 256$} on \mnist.}		(a) Test error rate: each row corresponds to one layer in the network.
		As a reference, the last row corresponds to a full model with
		parameters loaded from the corresponding checkpoint. The
		first column designates robustness of each layer w.r.t re-randomization
		and the rest of the columns designate re-initialization robustness at
		different checkpoints. The last column shows the performance of the
		final trained model (i.e. checkpoint-T) for reference.		(b-c) Weight distances: each cell in the heatmaps depicts the normalized
		$2$-norm (b) or $\infty$-norm (c) distance of trained parameter vectors
		to their initial value.
  }
  \label{fig:mnist-mlp-3x256}
\end{figure*}

We start by examining the robustness of fully-connected networks (\fcn{s}). A
\fcn{~$D\times H$} consists of $D$ fully connected layers each of output
dimension $H$ followed by a ReLU activation function. The additional final
layer is a linear multiclass predictor with one output per class.

We train an \fcn{~$3\times 256$} on \mnist{} and apply the re-initialization
and re-randomization analysis on the trained model. The results are shown in
Fig.~\ref{fig:mnist-mlp-3x256}(a). As expected, due to the intricate
dependency of the classification function on each of the layers,
re-randomizing any of the layers completely disintegrates the representation
and classification accuracy drops to the level of random guessing.  For
re-initialization, however, while the first layer is very sensitive the rest
of the layers are robust to re-initialization.

A plausible explanation for this could be attributed to that the increase
in gradient norms during back-propagation such that the bottom layers are
being updated more aggressively than the top ones. However, if this were the
case, we would expect a smoother transition instead of a sharp one at the
first layer. Furthermore, we measured how distant the weights of each layer are
from their initialization (``checkpoint-0'') using both 
the $2$-norm (divided by square root of the dimension) and the $\infty$-norm.
The results are shown in Fig.~\ref{fig:mnist-mlp-3x256} parts (b) and (c),
respectively.  We can see that robustness to re-initialization is not
plainly correlated with either of the distances. This suggests that there
might be something more intricate going on than simple gradient expansion.
We informally summarize the observations as follows,
\begin{quote}
  \emph{Over-capacitated deep networks trained with stochastic gradient descent
  have low-complexity by a self-restriction of the number of \critical layers.}
\end{quote}

Intuitively, if a subset of parameters can be re-initialized to the random
values at checkpoint-$0$ (which are independent of the training data), then the
the complexity of the model can be reduced. See the discussions in 
Section~\ref{sec:discussion} for more details.

\begin{figure*}\centering
  \begin{subfigure}{.43\linewidth}
    \begin{overpic}[width=\linewidth]{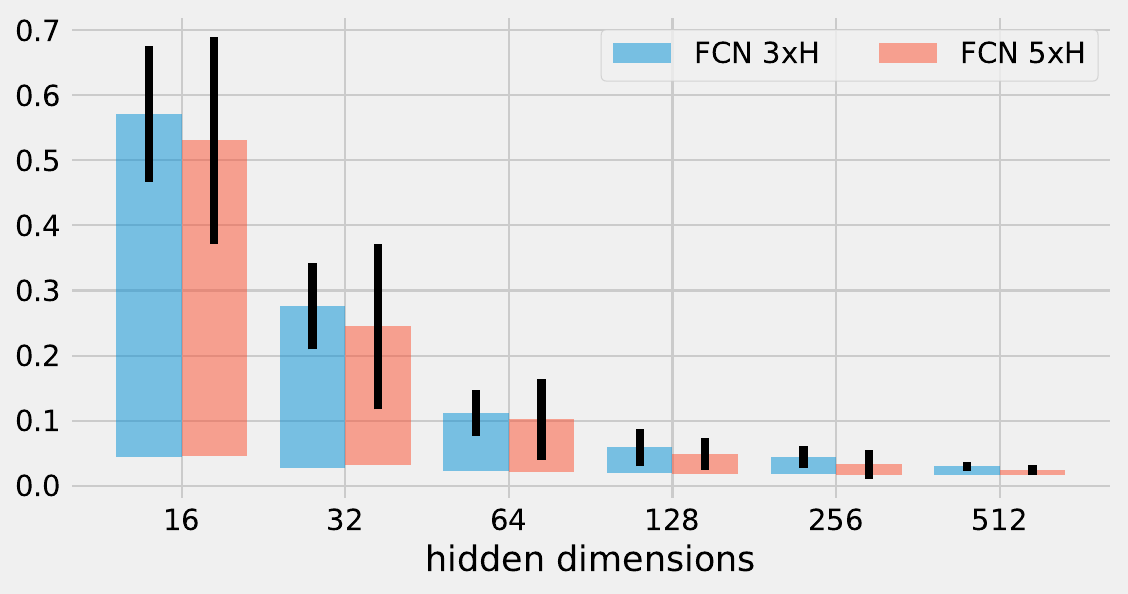}
      \put(-6,8){\rotatebox{90}{\scriptsize\textsf{Average test error}}}
    \end{overpic}
  \caption{MNIST}
  \end{subfigure}
  \begin{subfigure}{.43\linewidth}
  \includegraphics[width=\linewidth]{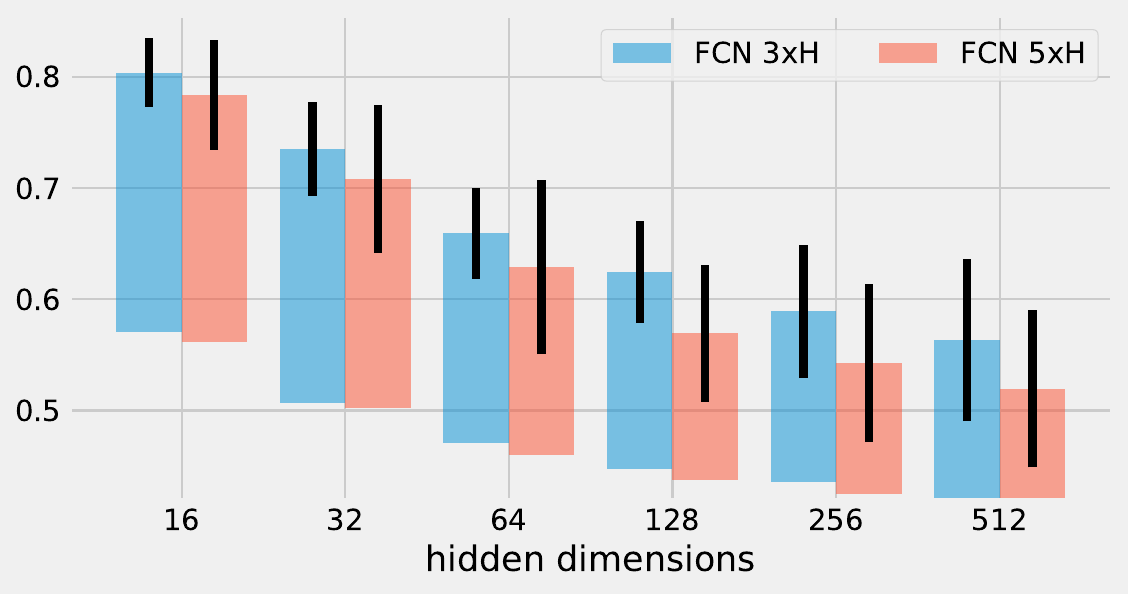}
  \caption{CIFAR10}
  \end{subfigure}
	\caption{\textbf{Average re-initialization robustness to checkpoint-0 for
	all layers but the first for \fcn{s}.} Each bar designates the difference
	in classification error between a model with one layer re-initialized (top
	of bar) and the same model without weight modification (bottom of bar).
	The error-bars designate one standard deviation obtained by running five
	experiments with different random initializations.}
  \label{fig:mlp-width-comparison}
\end{figure*}

We apply the same analysis framework to a large number of different \fcn{}
architectures to assess the influence of the network capacity and the task
complexity on the layer robustness.  In
Fig.~\ref{fig:mlp-width-comparison}(a), we compare the average
re-initialization robustness for all layers but the first with respect to
\fcn{s} $3\times H$ and $5\times H$ of varying hidden dimensions ($H$) on
\mnist. It is clear that the top layers become more robust as the hidden
dimension increases. We believe that it reflects the fact that wider
\fcn{s} have higher capacity. When the capacity is small, all layers
are vigil participants in representing the prediction function. As the
capacity increases, it suffices to use the bottom layer while the rest act
as random projections with non-linearities.

Similarly, Fig.~\ref{fig:mlp-width-comparison}(b) shows experiments on
\cifar{}, which has the same number of classes and comparable number of
training examples as \mnist yet it poses a more difficult learning task.
Similar traits are observed as the hidden dimensions increase, though not as
pronounced as in \mnist. Informally put, the difficulty of the learning task
seems to necessitate more diligence of the layers in forming accurate
predictors.

The empirical results of this section provide some evidence that deep
networks trained with SGD \emph{automatically} adjust their capacity.
When a large network is trained on an easy task, only a few layers seem
to be playing a critical role.

\subsection{Large Convolutional Networks}

\begin{figure}\centering
    \begin{subfigure}{.31\linewidth}
      \includegraphics[width=\linewidth]{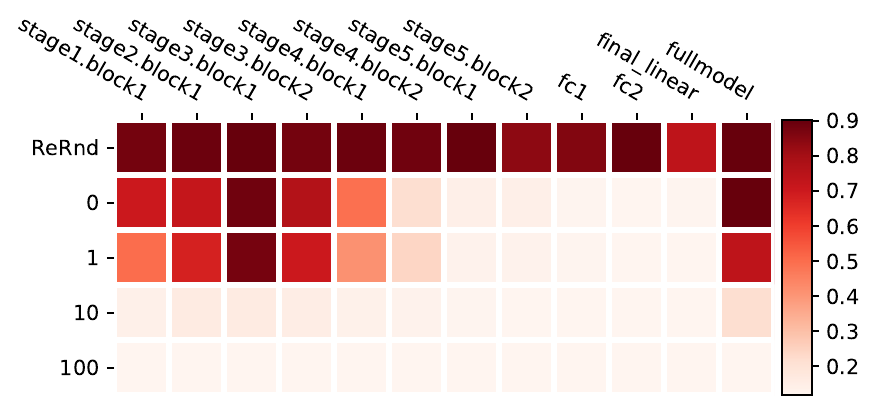}
      \caption{\vgg{-11}}
    \end{subfigure}\hspace{15pt}
    \begin{subfigure}{.55\linewidth}
      \includegraphics[width=\linewidth]{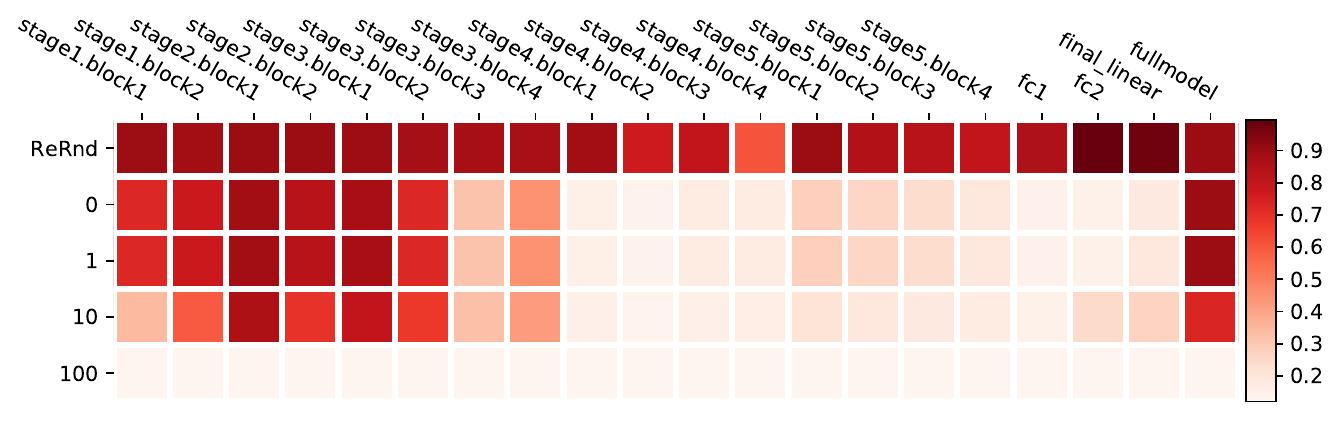}
      \caption{\vgg{-19}}
    \end{subfigure}
    \begin{subfigure}{.38\linewidth}
      \includegraphics[width=\linewidth]{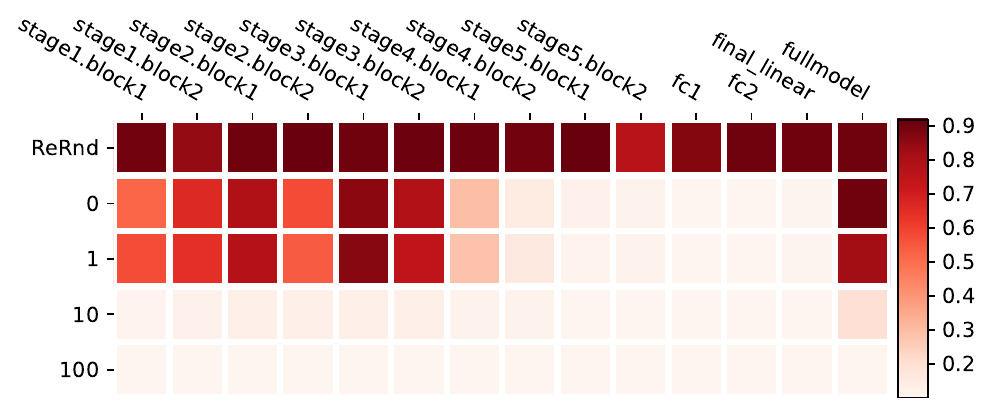}
      \caption{\vgg{-13}}
    \end{subfigure}\hspace{15pt}
    \begin{subfigure}{.47\linewidth}
      \includegraphics[width=\linewidth]{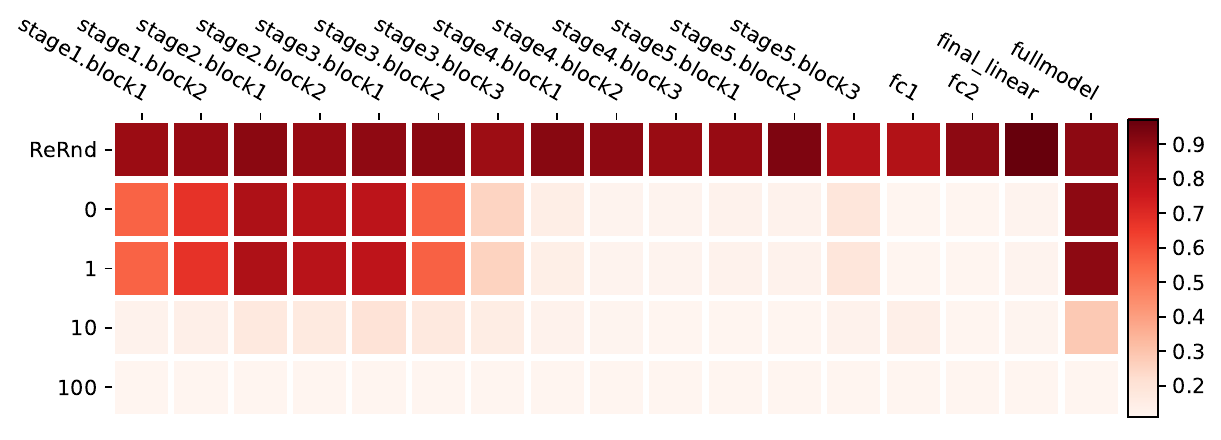}
      \caption{\vgg{-16}}
    \end{subfigure}
    \caption{\textbf{Robustness for \vgg{} networks on \cifar.}     Heatmaps use the same layout as in Fig.~\ref{fig:mnist-mlp-3x256} after
    being transposed to visualize the deeper architecture more effectively.}
    \label{fig:cifar10-vgg}
\end{figure}

\begin{figure}\centering
  \begin{subfigure}{.31\linewidth}
    \includegraphics[width=\linewidth]{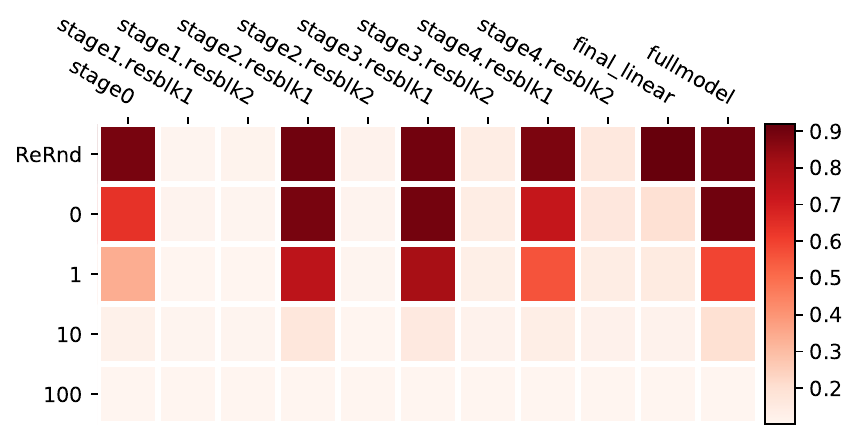}
    \caption{\resnet{-18}}
  \end{subfigure}\hspace{10pt}
  \begin{subfigure}{.55\linewidth}
    \includegraphics[width=\linewidth]{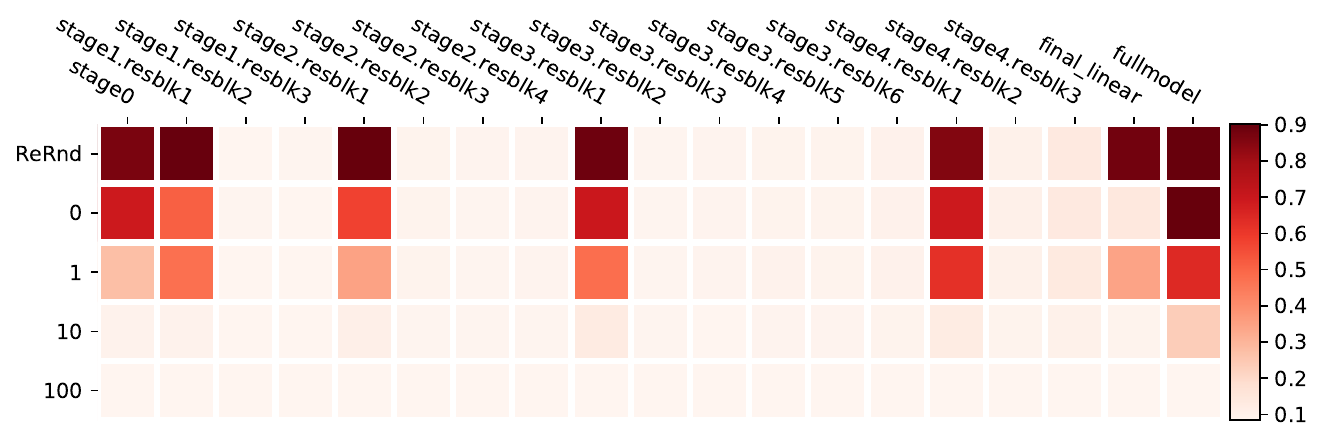}
    \caption{\resnet{-50}}
  \end{subfigure}\\
  \begin{subfigure}{.89\linewidth}
    \includegraphics[width=\linewidth]{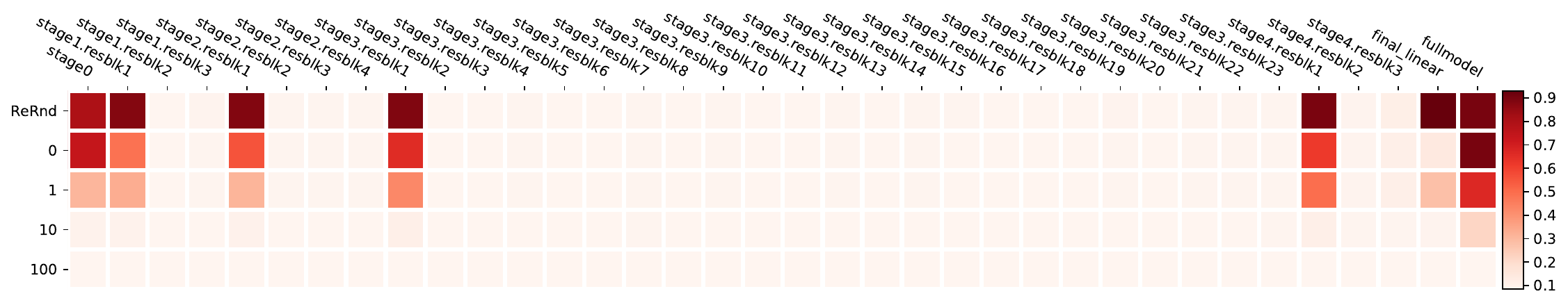}
    \caption{\resnet{-101}}
  \end{subfigure}
  \begin{subfigure}{.88\linewidth}
    \includegraphics[width=\linewidth]{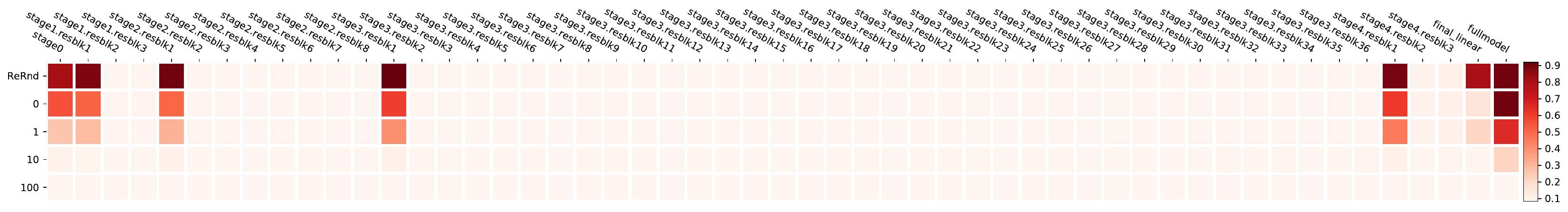}
    \caption{\resnet{-152}}
  \end{subfigure}
  \caption{\textbf{Robustness for residual blocks of
    \resnet{s} trained on \cifar.}}
  \label{fig:cifar10-resnet}
\end{figure}

In typical computer vision tasks beyond elementary problems such as \mnist,
densely connected \fcn{s} are significantly outperformed by convolutional
neural networks. \vgg{s}~\citep{simonyan2014very} and
\resnet{s}~\citep{he2016deep,he2016identity} are two widely used
convolutional network architectures. Fig.~\ref{fig:cifar10-vgg} and
Fig.~\ref{fig:cifar10-resnet} show the robustness results on \cifar{} with
the two architectures. Since the networks are much deeper than the
\fcn{s} of the previous section, we transpose the heatmaps and now a column
designates a layer. For \vgg{s}, more layers are sensitive to
re-initialization, yet the characteristics are similar to the observations
from the simple \fcn{s} on \mnist: bottom layers are evidently more
sensitive than the top layers to re-initialization.

\begin{figure}\centering
  \begin{subfigure}{.28\linewidth}
    \includegraphics[width=\linewidth]{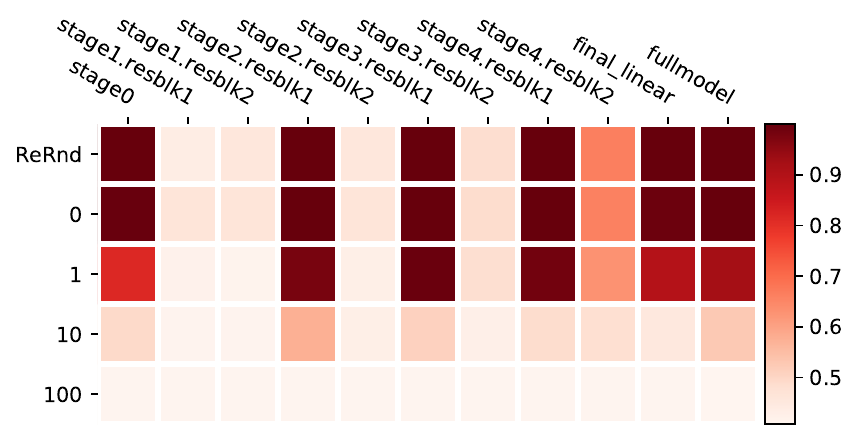}
    \caption{\resnet{-18}}
  \end{subfigure}\hspace{10pt}
  \begin{subfigure}{.52\linewidth}
    \includegraphics[width=\linewidth]{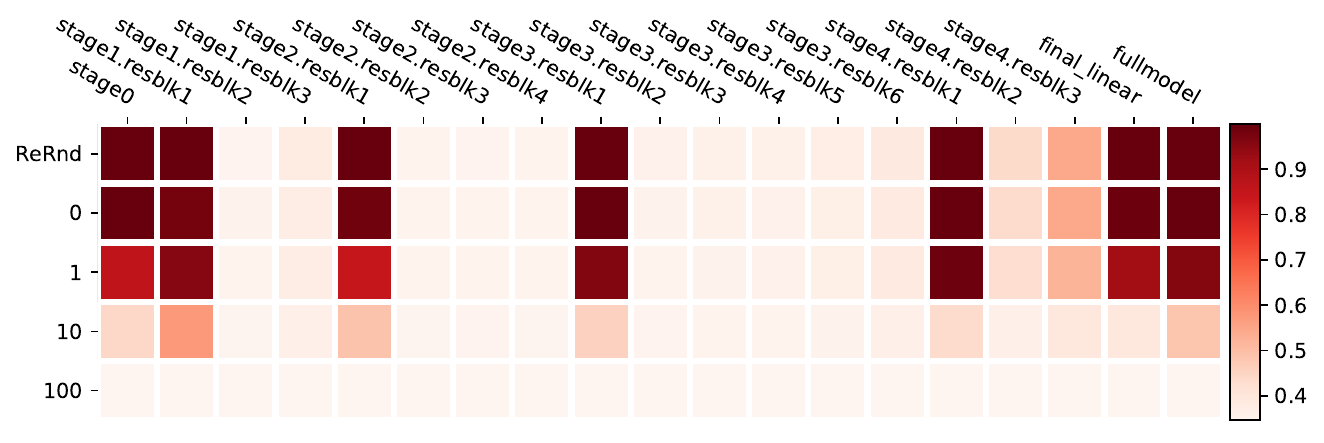}
    \caption{\resnet{-50}}
  \end{subfigure}
  \begin{subfigure}{.86\linewidth}
    \includegraphics[width=\linewidth]{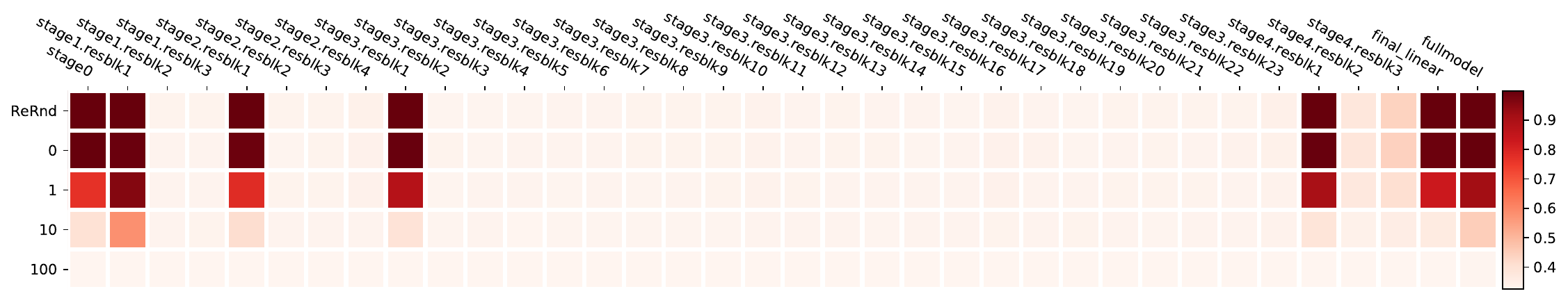}
    \caption{\resnet{-101}}
  \end{subfigure}
  \begin{subfigure}{.85\linewidth}
    \includegraphics[width=\linewidth]{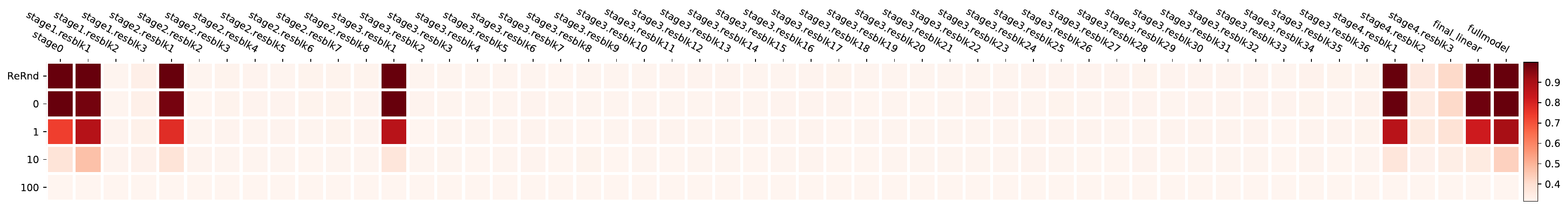}
    \caption{\resnet{-152}}
  \end{subfigure}
  \caption{\textbf{Robustness analysis for residual blocks of \resnet{s} trained
		on \imagenet.}}
  \label{fig:imagenet-resnet}
\end{figure}

The results for \resnet{s} in Fig.~\ref{fig:cifar10-resnet} are to be
considered together with results on \imagenet{} in
Fig.~\ref{fig:imagenet-resnet}. We found the robustness structure for
\resnet{s} to be more interesting for the reasons below.

\paragraph{\resnet{s} re-distribute \critical layers.} Unlike \fcn{} and
\vgg{} networks for which the \critical layers are at the bottom of the
network, \resnet{s} sprinkles critical layers throughout the entire depth.
To better understand the patterns, let us briefly recap \resnet{}'s
architecture. In practice, a \resnet{} is divided into ``stages''. At the
bottom, there is a pre-processing stage (\texttt{stage0}) with vanilla
convolutional layers. It is followed by a few (typically 4) residual stages
consisting of multiple residual blocks, and lastly a global average pooling
operator followed by a linear classifier (\texttt{final\_linear}). The image
size halves and the number of convolution channels doubles from each
residual stage to the next.\footnote{There are more subtle details which we
omit, especially at \texttt{stage1} that depend on the input
size, whether residual blocks contain a bottleneck, etc.}. As a result,
while most of the residual blocks have \emph{identity} skip
connections, the first block of each stage (\texttt{stage*.resblk1}), which
is connected to the last block of the previous stage, has a
\emph{non-identity} skip connection due to different input-output shapes.
Fig.~\ref{fig:resblk-illustration} in the Appendix illustrates the two types
of residual blocks. In our robustness analysis, we can interpret each stage
of a \resnet{} as a sub-network, with characteristics of layer robustness
\emph{within} each stage similar to \vgg{s} or \fcn{s}.

\paragraph{Residual blocks show robustness to re-randomization.} Among the
layers that are robust to re-initialization, if the layer is a residual
block, it is also robust to re-randomization, which stands in contrast to
the \texttt{final\_linear} layer. This could be potentially attributed to
that the responses for identity skip connections attain larger values
than those of the residual branches. Thus, when summed together residual
branches played a less significant role. It is known from prior
research~\citep{veit2016residual} that residual blocks in a \resnet{} can be
removed without substantially hurting accuracy. Our experiments have a
different focus as we study robustness in the light of the interplay between
model capacity and the difficulty of the learning task. In particular,
comparing the results on the two different datasets, especially on smaller
\resnet{s} (e.g. \resnet{-18}), many residual blocks with real identity skip
connection also become sensitive in the more difficult \imagenet{} task.

\section{Joint Robustness}
\label{sec:joint-robustness}
Empirical results we presented thus far focus on individual layer
robustness. We next explore \emph{joint robustness} of multiple layers through
\emph{simultaneous} re-initialization or re-randomization.

\begin{figure}\centering
  \begin{subfigure}{.85\linewidth}
    \includegraphics[width=\linewidth]{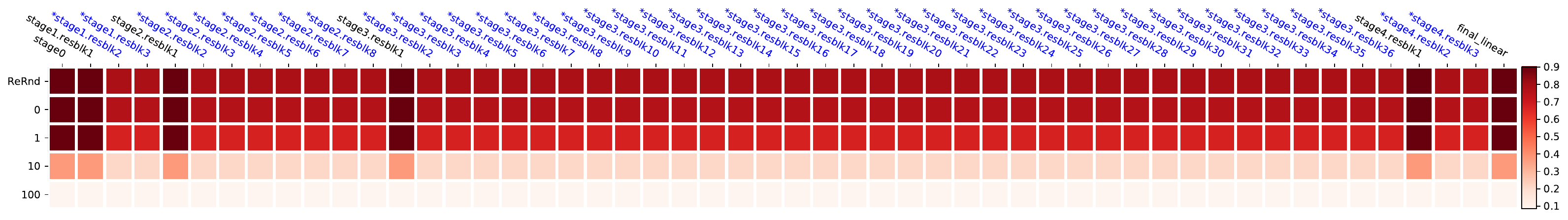}\vskip-5pt
    \caption{\scriptsize\resnet{152}: \textrm{resblk2, 3, }\ldots}
  \end{subfigure}
  \begin{subfigure}{.85\linewidth}
    \includegraphics[width=\linewidth]{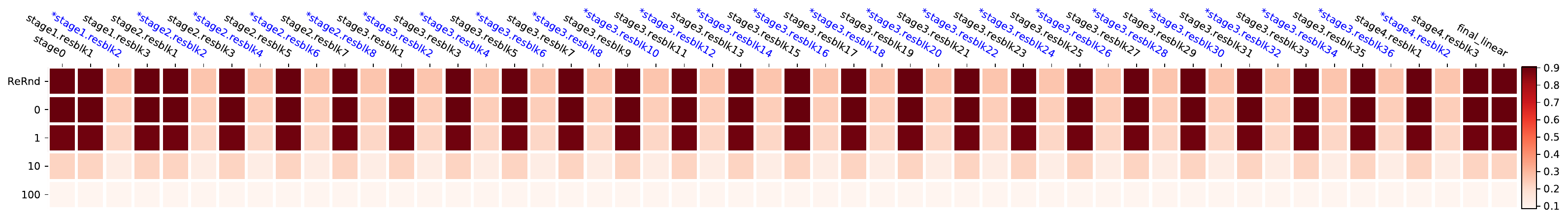}
    \caption{\scriptsize\resnet{152}: every second \textrm{resblk}}
  \end{subfigure}
   \caption{\textbf{Joint robustness of \resnet{152} on \cifar}.
   Jointly re-initialized/re-randomized group of layers are indicated with
   \texttt{*} over the layer name and are also printed in \textcolor{blue}{blue}
	 for easy identification.}
  \label{fig:bulk-resnet-body}
\end{figure}

We divide the layers into two groups and perform robustness experiments with
each group. For \resnet{s}, we put all but the first residual blocks of all
stages into one group and jointly re-initialize or re-randomize all the
layers in the group.  Fig.~\ref{fig:bulk-resnet-body}(a) demonstrates that
even though each of these layers are all individually \robust, as a group
they are \emph{not} jointly \robust. However, a different grouping scheme
shown in Fig.~\ref{fig:bulk-resnet-body}(b) demonstrates that robustness can
be significantly improved when jointly resetting about half of the layers
for this \resnet{} architecture. See Appendix~\ref{app:joint-robustness} for
more details.

Note that SGD fits the model with the aforementioned jointly robust
structure without explicit constraints. We next examine if explicit
constraints could improve joint robustness. Concretely, we experiment with
two approaches: (i) We freeze layers at their random initialization and
refrain from training all frozen layers; (ii) We remove layers from the
network. Both operations are applied to the groups described above excluding
again the first residual block of each stage.

\begin{table}
  \caption{\normalsize \textbf{Error rates(\%) on \cifar{} (top) and
  \imagenet{} (bottom).} Each row reports the performance of a full model, average
  individual layer robustness to re-initialization (mean$\pm$std), partially trained
  models with a subset of the layers frozen to their initial values, and
  partially trained model after removing a subset of the layers. Individual layer
  robustness is measured \emph{separately} and averaged across all but the first
  residual blocks of all stages. Layer-freezing and layer-removal are \emph{jointly}
	applied to the same set of residual blocks.}
  \label{tab:freeze-n-remove}
  \centering\vskip4pt
  \begin{tabular}{p{.1em}lcccc}
  \toprule
  & \multirow{1}{*}{Arch} & Full Model & Individual Layer Robustness & Layers Frozen & Layers Removed \\
  &      & & (Average) & (Jointly) & (Jointly) \\
  \midrule
  \parbox[t]{2mm}{\multirow{3}{*}{\rotatebox[origin=c]{90}{\scriptsize\cifar}}}
& \resnet{50} & 8.40 & 9.77$\pm$1.38 & 11.74 & 9.23 \\ & \resnet{101} & 8.53 & 8.87$\pm$0.50 & 9.21 & 9.23 \\ & \resnet{152} & 8.54 & 8.74$\pm$0.39 & 9.17 & 9.23 \\ \midrule
  \parbox[t]{2mm}{\multirow{3}{*}{\rotatebox[origin=c]{90}{\scriptsize\imagenet}}}
& \resnet{50} & 34.74 & 38.54$\pm$5.36 & 44.36 & 41.50 \\ & \resnet{101} & 32.78 & 33.84$\pm$2.10 & 36.03 & 41.50 \\ & \resnet{152} & 31.74 & 32.42$\pm$1.55 & 35.75 & 41.50 \\   \bottomrule
  \end{tabular}
\end{table}
The results are given in Table~\ref{tab:freeze-n-remove}. When we freeze
layers, the resulting error is higher than that of the average individual layer
robustness measured in a normally trained model. However, the gap is much smaller than
when directly assessing the joint robustness. Moreover, on \cifar{}, we find
that similar performance can be achieved even if we entirely remove those
layers from the network. In contrast, for \imagenet{} layer removal results
in a significant drop in performance. In this case, random projections followed
by non-linear activations conducted by frozen layers deem necessary to
maintain accuracy.

\begin{figure}
  \includegraphics[width=\linewidth]{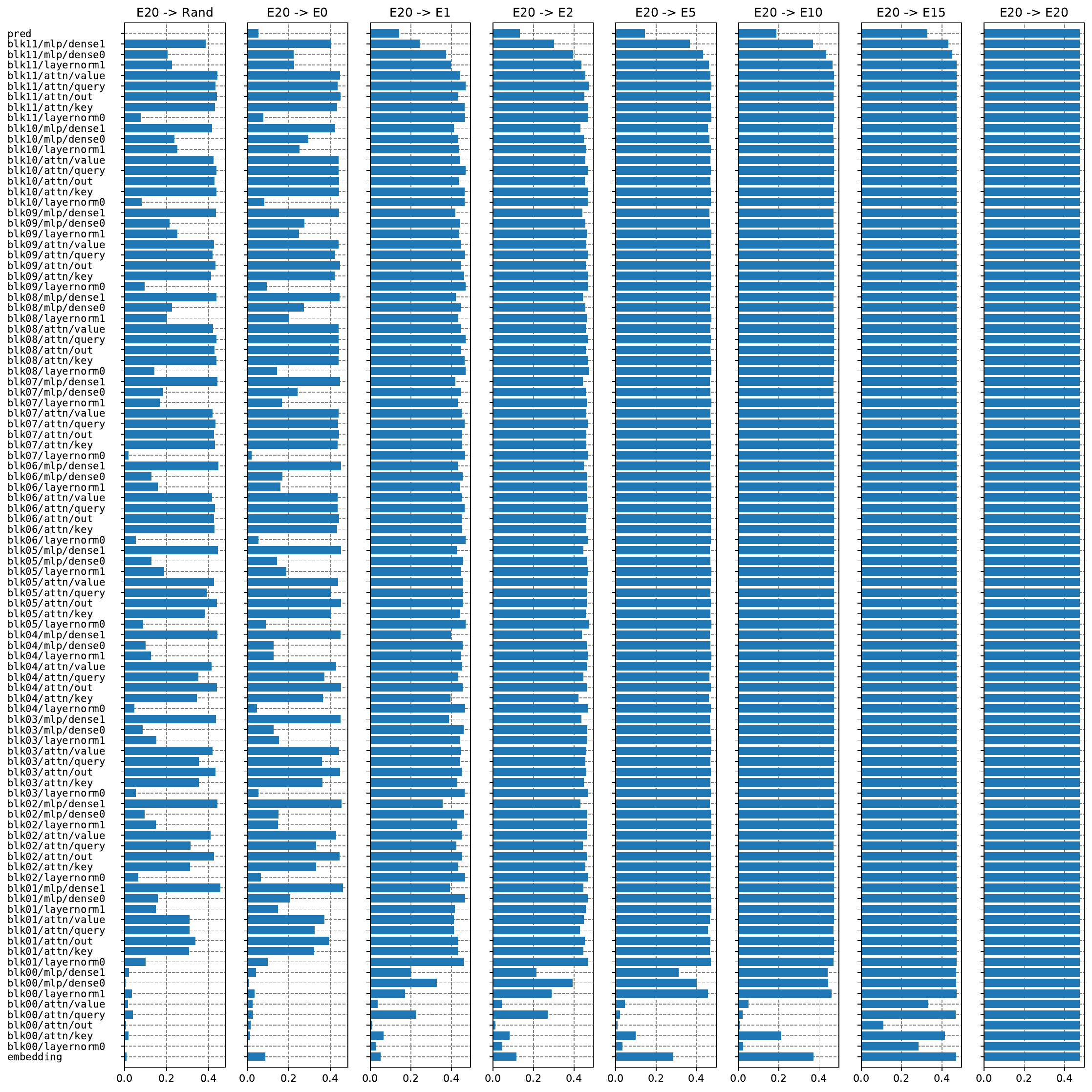}
  \caption{\textbf{Robustness results of Transformer-based neural language model trained on LM1B}. 
  The robustness is measured with average per-token accuracy on the validation set. The $i$-th row
  of each block shows the model performance after re-initializing the $i$-th layer to the checkpoint
  designated on the block title. The first block shows re-randomization results. This model is trained
  for 20 epochs, so the last block (E20 $\rightarrow$ E20) shows the performance of the un-modified full model
  as a reference.}
  \label{fig:lm1b}
\end{figure}

\section{Other Architectures and Domains}
\label{sec:other-domains}

In this section, we study the generalizability of the layer robustness phenomenon by extending the evaluation to 
a different domain (language modeling) and different model families (convolution-free architectures). We
find the results corroborate the key observations made earlier in this paper.

\subsection{Transformer Based Neural Language Models}

We consider a 12-layer decoder-only~\citep{liu2018generating} Transformer~\citep{vaswani2017attention} based 
neural language model, more specifically, the decoder-only version of the T5-Base~\citep{2020t5} model. It
has 112,242,480 parameters. We train it on the LM1B~\citep{DBLP:journals/corr/ChelbaMSGBK13} dataset, which
is a popular benchmark corpus for language modeling with 30,301,028 training examples (around one billion
words). We use the SentencePiece\footnote{\url{https://github.com/google/sentencepiece}.} tokenizer with a
vocabulary size of 30,000. We train the model with Adam optimizer~\citep{kingma2014adam} for 20 epochs.

We then measure the robustness of this model on the validation set with the average per-token accuracy.
Figure~\ref{fig:lm1b} show the results for re-randomization, and re-initialization to epoch 0, 1, 2, 5,
10, 15 and 20, respectively. The results for re-randomization and re-initialization to epoch-0 look very
similar, except for the \texttt{embedding} and the \texttt{pred} layers. Note the \texttt{pred} layer
performs a 30,000 way classification to predict the next token. This is consistent with our earlier observation 
comparing CIFAR-10 and ImageNet that the final prediction layer is only robust when the number of classes
is small. In a language model, the \texttt{embedding} layer maps each of the 30,000 input tokens to a
dense embedding vector, and in some architectures it simply shares weights with the \texttt{pred} layer.

The overall trend that higher layers are more robust than lower layers still holds. But we also see
some patterns unique to the transformer architecture. For example, the \emph{layer normalization}~\citep{ba2016layer}
layers are generally not robust. The first dense layer in each of the MLP block is also sensitive 
to re-initialization or re-randomization. On the other hand, the second dense layer in each MLP block,
as well as all the components in the attention blocks are generally robust (except for \texttt{block00/*}).

Another interesting difference comparing to the vision experiments is that many of the unrobust layers
become robust after only one epoch of training. This is likely due the much larger training set sizes
in the text domain.

\begin{figure}
  \centering
  \includegraphics[width=\linewidth]{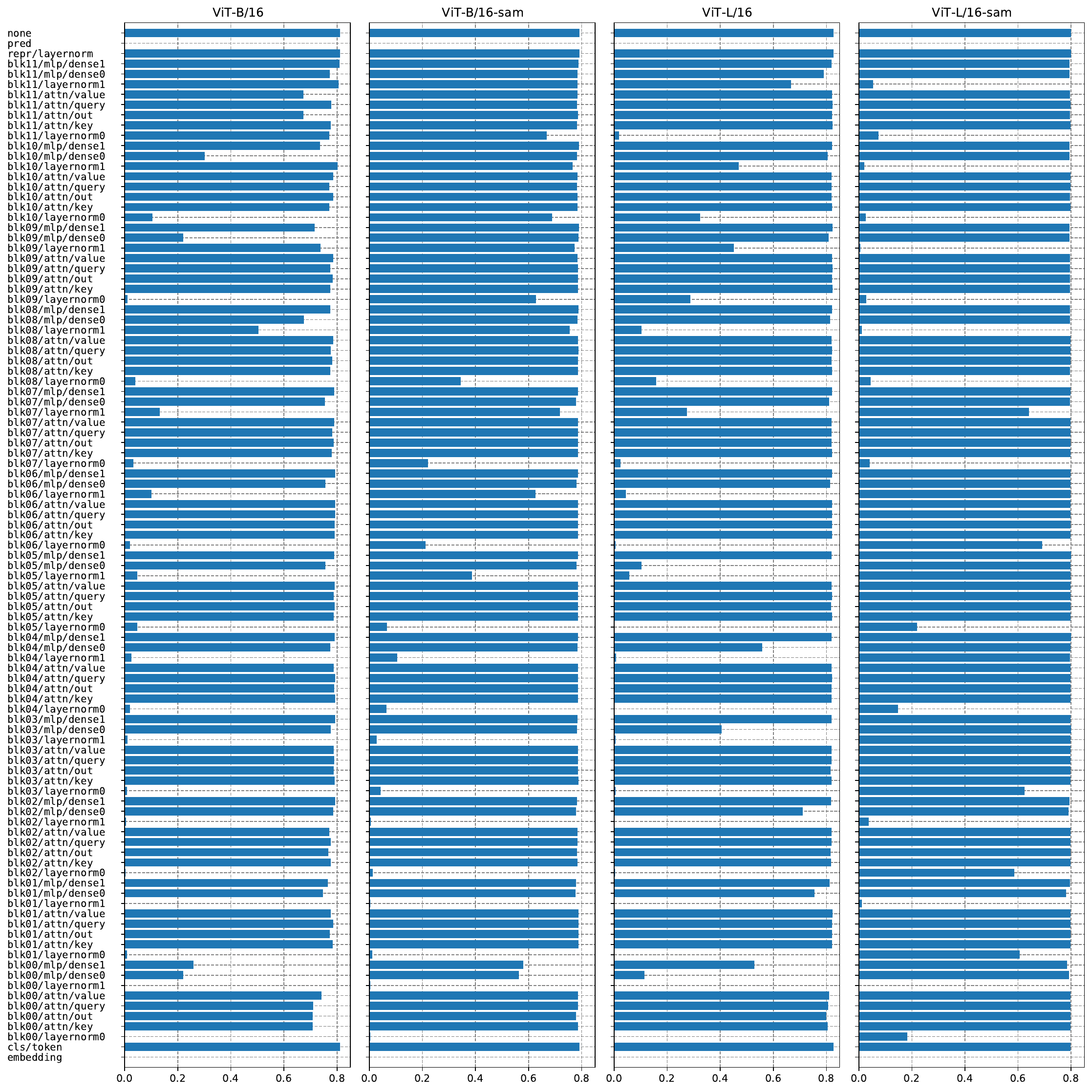}
  \caption{\textbf{Re-randomization robustness for Vision Transformers (ViTs),
  measured by validation accuracy on \imagenet-1k}. 
  The first row (\texttt{none}) is the full model performance without re-randomization.}
  \label{fig:rerand-vit}
\end{figure}

\begin{figure}
  \centering
  \includegraphics[width=\linewidth]{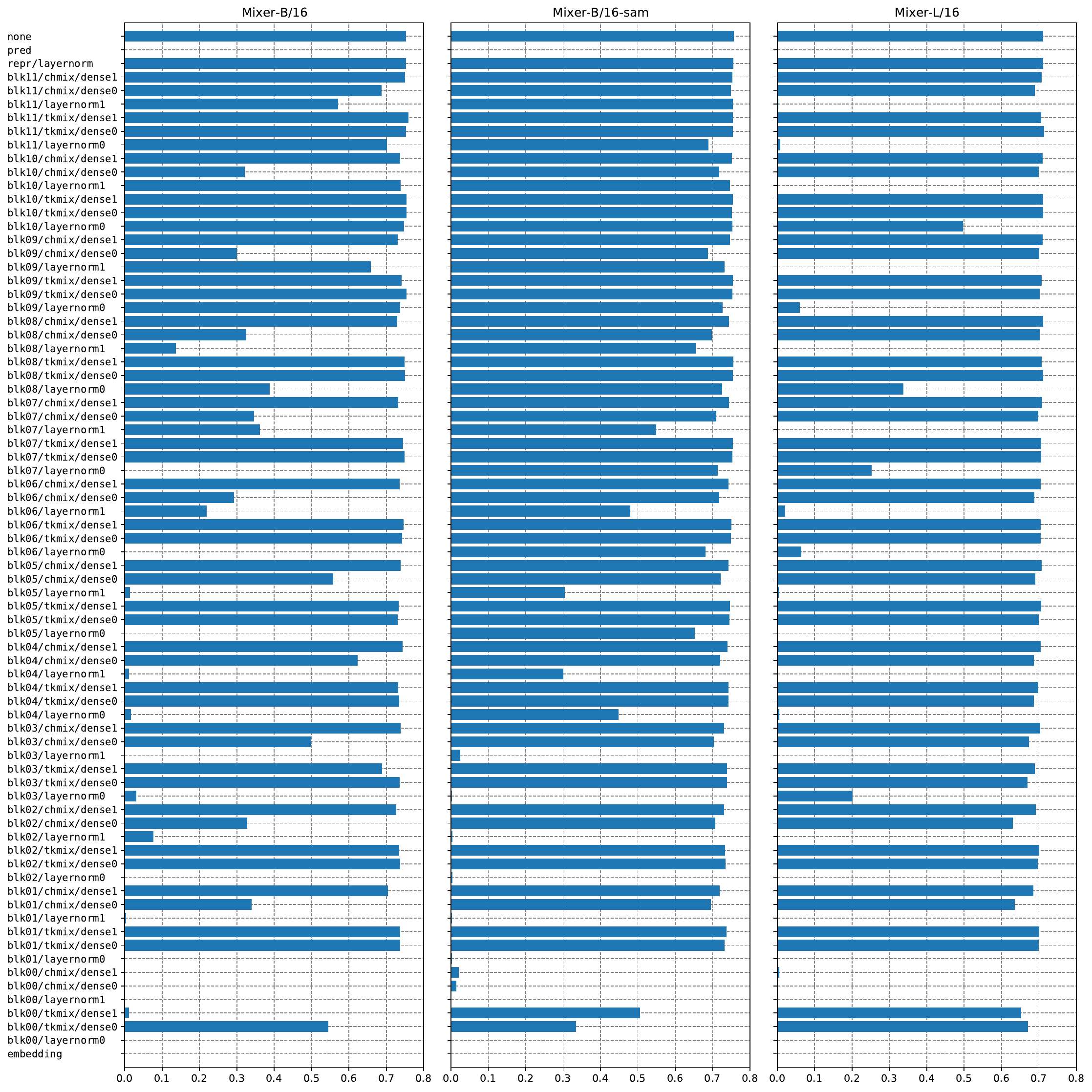}
  \caption{\textbf{Re-randomization robustness for MLP-Mixers,
  measured by validation accuracy on \imagenet-1k}. 
  The first row (\texttt{none}) is the full model performance without re-randomization.}
  \label{fig:rerand-mixer}
\end{figure}

\subsection{Convolution-Free Architectures for Computer Vision}

In this section, we provide preliminary studies for two new neural network architectures
in computer vision that achieve the state-of-the-art performance in image classification benchmarks
\emph{without} using convolution layers. In particular, we consider the attention~\citep{vaswani2017attention}
based Vision Transformers~\citep[ViTs,][]{dosovitskiy2020vit} and the MLP based MLP-Mixers~\citep{tolstikhin2021mlp}.
We use the publicly released pre-trained checkpoints\footnote{\url{https://github.com/google-research/vision_transformer}.}
that are trained for \imagenet. Since two variants of \imagenet datasets~\citep{deng2009imagenet} were used when training those models, in
this subsection, we will spell out the variants as \imagenet-21k --- the larger dataset containing 21,000 classes
and 14M training images, and \imagenet-1k --- the smaller dataset containing 1,000 classes and 1.2M training images.
In particular, \imagenet-1k is the same as the ``standard'' \imagenet dataset used in the rest of this paper. 

We run robustness evaluation on the \imagenet-1k validation set. We use checkpoints that can be directly
evaluated on \imagenet-1k. In particular, for ViTs, we have access to checkpoints that are pre-trained on
\imagenet-21k and then finetuned on \imagenet-1k. We evaluate two variants: ViT-B/16 (86,567,656 parameters) 
and ViT-L/16 (304,326,632 parameters). 
For MLP-Mixers, we have access to checkpoints that are directly trained on \imagenet-1k. 
We also evaluate two variants: Mixer-B/16 (59,880,472 parameters) and Mixer-L/16 (208,196,168 parameters). 
Additionally, several new checkpoints trained
with the \emph{Sharpness-Aware Minimization}~\citep[SAM,][]{foret2020sharpness} became available recently.
We include ViT-B/16-sam, ViT-L/16-sam and Mixer-B/16- in the comparisons (Mixer-L/16-sam checkpoint is
not available). All the SAM optimized models are directly trained on \imagenet-1k.
Since we do not have access to the original training pipeline and earlier checkpoints, we only perform
re-randomization tests on those models.

Figure~\ref{fig:rerand-vit} shows the results for ViTs. The ViT architectures are based on the Transformer
models for text processing, and we observe the layer robustness patterns are similar to the results on
Transformer based language models in Figure~\ref{fig:lm1b}: the attention related layers and the 
second dense layer of MLP blocks (\texttt{mlp/dense1}) are generally robust, especially for higher up blocks.
We do find more layers (e.g. \texttt{mlp/dense0}) robust here. Comparing among the variants, we find that
the larger ViT-L/16 model is generally more robust than the smaller ViT-B/16 model for most layers, except
for some \texttt{layernorm} layers the reverse is true. This makes the gap between the robust and non-robust
layers more pronounced in the larger model. Comparing each ViT variant with its SAM-optimized version, we
see that the SAM optimizer generally improves the robustness for ViT-B/16, and for ViT-L/16 it has an additional
effect of making some higher-up \texttt{layernorm} layers even less robust, further enlarging the gap.
Figure~\ref{fig:rerand-mixer} shows the results for MLP-Mixers. The overall patterns are similar to that
of the ViTs.

\section{Connections to Other Notions of Robustness}
\label{sec:other-robustness}
Layer robustness to
re-initialization and re-randomization can be related to other notions of
robustness in deep learning. For example, \emph{flatness} refers to
robustness to \emph{local} perturbations of the network's parameters close
to a converged model, and is extensively discussed in the
context of generalization \citep{hochreiter1997flat, chaudhari2016entropy,
keskar2016large, smith2018bayesian, 3694}. For a fixed layer, our notion of
robustness to re-initialization is restricted to the training
trajectory, which could potentially take the form of a \emph{non-local}
perturbation.  Robustness to re-randomization allows for larger
variances of perturbations of the trained parameters. As our study shows,
robustness seems to be layer-dependent, thus analyzing layers individually
for specific network architectures enables us to obtain more refined insights
into robustness.

In contrast, \emph{adversarial} robustness~\citep{szegedy2013intriguing}
focuses on robustness to perturbations of the input. In particular, it was
found that deep networks are sensitive to small adversarial perturbations
which yield prediction shifts to arbitrary classes. A large number of
defense and attack algorithms have been proposed in recent years. Here we
briefly discuss the connection to adversarial robustness. Take a standard
\resnet{} with $S$ stages of $(B_1,\ldots,B_S)$ residual blocks in each
stage. At test time, we turn it into a randomized classifier by selecting at
random a subset of $s\in\{0,1,\ldots,S\}$ stages, and replacing at random a
residual block from each of the selected stages with one of the $r$
pre-initialized weights of its layer. We keep $r$ pre-allocated weights for
each residual block instead of re-sampling at random on each evaluation
call, primarily to reduce computation during the testing.

From the robustness analysis of previous sections, we expect randomized
classifiers to exhibit only a small drop in \emph{average} performance.
However, at the individual example level, randomization of the network's
outputs would make it harder for an attacker to generate adversarial
examples.  We evaluate the adversarial robustness against a weak FGSM
\citep{goodfellow6572explaining} attack and a strong PGD
\citep{madry2017towards} attack. The results in Table~\ref{tab:adversarial}
show that compared to the baseline (identical model without output
randomization), output randomization significantly
improves robustness to weak FGSM attacks. The performances under strong PGD
attack drops sharply yet it is still an order of magnitude better than the
baseline. The results relate layer robustness to adversarial robustness, 
but it does not imply randomization via robust layers provides strong 
adversarial robustness, especially under attacks that are 
specifically designed with such randomization in mind~\citep{athalye2018obfuscated}.

To recap, layer robustness can guard a trained model from attack by
injecting randomization. However, robustness per se does not provide
sufficient defense against strong attacks. More sophisticated attacks that
explicitly deal with non-deterministic classifiers could completely
render the approach unusable.

\begin{table}\small

\caption{\normalsize \textbf{Accuracy(\%) of various model configurations on
	clean \cifar{} test set, under weak (FGSM), and strong (PGD) adversarial
	attack.} Adversarial attacks are evaluated on a subset of 1000 test
	examples. Every experiment is repeated 5 times and the average performance
	is reported. The hyperparameters $r$ and $s$ in model configurations
	correspond to the number of random weights pre-set for each residual
	block, and the number of stages that are re-randomized during each
	inference step. Here $\mbox{ResN}(4^2)$ designates an architecture with
	two stages, each stage of four residual blocks. Similarly, $\mbox{ResN}(4^4)$
	has four stages each with four residual blocks.}
\label{tab:adversarial}
\centering\vskip4pt
  \begin{tabular}{llcccc}\toprule
   \multicolumn{2}{c}{Model Configuration}    & Clean & FGSM & PGD \\
 \midrule
		\multirow{3}{*}{$\mbox{ResN}(4^2)$}
  & baseline & $91.05\pm0.00$ & $12.75\pm0.04$ & $0.33\pm0.16$\\
  & r=4,s=1 & $89.45\pm0.13$ & $69.85\pm1.60$ &  $6.71\pm0.37$\\
  & r=4,s=2 & $87.70\pm0.25$ & $71.18\pm0.49$ &  $9.65\pm0.27$\\
  \midrule
		\multirow{4}{*}{$\mbox{ResN}(4^4)$}
  & baseline & $90.08\pm0.00$ & $8.45\pm0.00$ &  $0.00\pm0.00$\\
  & r=4,s=1 & $89.64\pm0.12$ & $62.76\pm1.09$ &  $2.60\pm0.26$\\
  & r=4,s=2 & $89.13\pm0.13$ & $67.20\pm0.63$ &  $3.56\pm0.48$\\
  & r=4,s=4 & $88.24\pm0.18$ & $69.09\pm1.59$ &  $5.60\pm0.53$\\
  \bottomrule
  \end{tabular}
\end{table}

\section{Discussion} \label{sec:discussion}
Excessive overparameterization of modern neural network architectures
renders conventional generalization bounds based on capacity
estimation of the entire hypothesis space unusable. Alternative approaches try to
identify nice properties such as bounds on the parameter norms of the models
trained with specific algorithms (e.g. SGD) on well behaved data, and derive
tighter generalization bounds based on those properties. Our experiments
provide useful evidence for deriving tighter generalization bounds via fine
grained analysis of layer behaviors.

For example, \citet{chatterji2019intriguing} formulated a notion of 
\emph{module criticality} based on our observation of the dichotomy between
critical and robust neural network layers, and derived PAC-Bayes generalization
bounds. We briefly present their theoretical results below.

\begin{definition}[Module and Network Criticality~\citep{chatterji2019intriguing}]
  Given an $\epsilon>0$ and network $f_\Theta$, we define the module criticality
  for module $i$ as follows:
  \begin{equation}
    \mu_{i,\epsilon}(f_\Theta) = \min_{0\leq \alpha_i, \sigma_i\leq 1}
    \left\{ 
      \frac{\alpha_i^2\|\theta_i^F - \theta_i^0\|_{\text{Fr}}^2}{\sigma_i^2}:
      \mathbb{E}_{u\sim \mathcal{N}(0,\sigma_i^2)} 
      [ \mathcal{L}_S(f_{\theta_i^\alpha + u,\Theta_{-i}^F}) ] \leq \epsilon
    \right\},
    \label{eq:criticality}
  \end{equation}
  We also define the network criticality as the sum of the module criticality over
  modules of the network:
  \begin{equation}
    \mu_\epsilon(f_\Theta) = \sum_{i=1}^d \mu_{i,\epsilon}(f_\Theta).
  \end{equation}
\end{definition}
Here $\mathcal{L}_S$ denotes the zero-one training loss. $\theta_i^0,\theta_i^F$ 
indicate the randomly initialized and final trained
value of the weight matrix of module $i$. $\theta_i^\alpha = (1-\alpha)\theta_i^0 + \alpha\theta_i^F$
is a convex combination of the two. $f_{\theta_i^\alpha,\Theta_{-i}^F}$ 
is the final trained neural network where the weight of its $i$-th module is replaced
with $\theta_i^\alpha$. Intuitively, a \emph{robust} layer could satisfy the condition in
\eqref{eq:criticality} with near-zero $\alpha$, therefore has a low criticality value.

\begin{theorem}[\citet{chatterji2019intriguing}]
  For any data distribution $D$, number of samples $m\in \mathbb{N}$, for any 
  $0<\delta <1$, for any $0<\sigma_i\leq 1$ and any $0\leq \alpha_i\leq 1$, with
  probability $1-\delta$ over the choice of the training set $S_m\sim D$ the following
  generalization bound holds:
  \begin{equation}
    \mathbb{E}_U[\mathcal{L}_D(f_{\Theta^\alpha + U})] \leq 
    \mathbb{E}_U[\mathcal{L}_S(f_{\Theta^\alpha + U})] + 
    \sqrt{\frac{\frac{1}{4}\sum_{i=1}^d k_i \log \left( 1 + 
        \frac{\alpha_i^2\|\theta_i^F - \theta_i^0\|_{\text{Fr}}^2}{k_i\sigma_i^2}
     \right) + \log(\frac{m}{\delta}) + \tilde{\mathcal{O}}(1)}{m-1}},
  \end{equation}
  where $k_i$ is the number of parameters in module $i$. $\Theta^\alpha$ indicate the
  network with the weights of each module $i$ replaced with $\theta_i^{\alpha_i}$, where
  $\alpha=(\alpha_i)_{i=1}^d$.
\end{theorem}

This PAC-Bayes bound characterize perturbed networks. Corollaries that characterize
the original network and that provides deterministic generalization can be found in 
Corollary~3.3 and Appendix~B of \citet{chatterji2019intriguing}. The existence of 
robust layers controls complexity terms in the bounds, providing alternative explanation
of the generalization power of large overparameterized neural networks. It is shown that 
this provides more faithful ranking of the generalization power of neural networks than 
many previous complexity measures \citep[][Section~4]{chatterji2019intriguing}.

To recap the paper, we empirically investigated the functional structure on
a layer-by-layer basis of overparameterized deep models for a wide variety
of models for image classification. We introduced the notions of
re-initialization and re-randomization robustness. Using these notions we
provided evidence for the heterogeneous nature of layers, which can be
categorized into either \robust or \critical. Resetting the \robust layers
to their initial value has negligible effect on the model's performance. Our
empirical results give further evidence that mere parameter counting or norm
accounting is too coarse in studying generalization of deep models.
Moreover, optimization landscape based analysis is better performed
respecting the network architectures due to the heterogeneous nature of
different layers. Our empirical work gives rise to several theoretical
questions. We conclude with a short list which is by no means comprehensive
of potential directions for future research motivated by our results.

\begin{description}

	\item[Lyapunov Function for Deep Architectures.] Bregman divergences over
		the {\em entire} set of parameters constitute the tool of choice in
		analyzing convergence and regret bounds for convex and quasi-convex
		models. Our results indicate that parameters of a network do {\em not}
		form a monolithic set. Devising novel {\em composite} divergences for
		measuring the progression of learning process is thus deemed useful.

	\item[Formation of \robust and \critical Layers.] As shown, some layers of
		a deep networks become robust to reset while other layers deem to be
		critical. This structural symmetry breaking could also shed light on
		the role of initialization for deep learning.

	\item[Hybrid Algorithms.] Our study underscores the potential of hybrid
		networks of mixed learned and random parameters. Random feature maps for
		deep learning were studied for fully random
		representations~\citep{rahimi2008random, rahimi2009weighted,
		daniely2016toward}. A potentially high impact direction is devising
		learning algorithms for building hybrid learned-random networks.

\end{description}

\acks{We would like to thank David Grangier, Lechao Xiao, Kunal Talwar,
Hanie Sedghi, and Omar Rivasplata for helpful discussions and feedback.}

\clearpage\appendix

\section{Details of Experimental Setup}
\label{app:exp-details}

Our empirical study is based on the \mnist, \cifar{} and the ILSVRC 2012
\imagenet{} datasets. Stochastic Gradient Descent (SGD) with a momentum of
0.9 is used to minimize the multi-class cross entropy loss. Each model is
trained for 100 epochs, using a stage-wise constant learning rate scheduling
with a multiplicative factor of 0.2 on epoch 30, 60 and 90. Batch size of
128 is used, except for \resnet{s} with more than 50 layers on \imagenet,
where batch size of 64 is used due to device memory constraints.

\begin{figure}[b]
  \begin{subfigure}{.48\linewidth}
    \includegraphics[width=\linewidth]{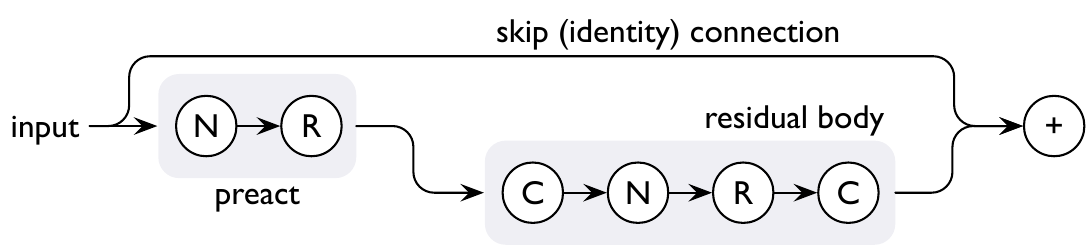}
    \caption{Residual block}
  \end{subfigure}\hfill
  \begin{subfigure}{.48\linewidth}
    \includegraphics[width=\linewidth]{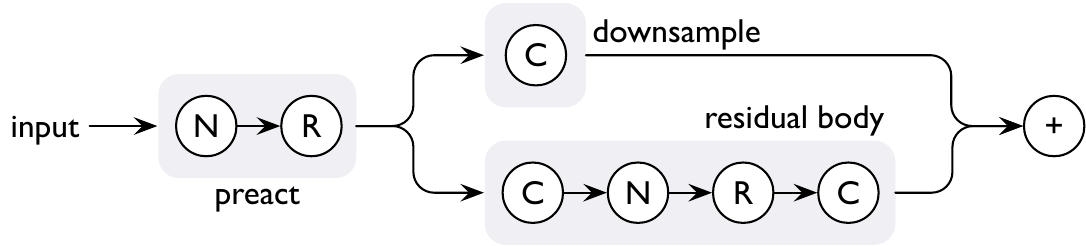}
    \caption{Residual block with downsampling}
  \end{subfigure}
  \caption{\textbf{Illustration of residual blocks (from \resnet{s} V2) with and without a downsampling skip branch.} \textsf{C}, \textsf{N} and \textsf{R} stand for convolution, (batch) normalization and ReLU activation, respectively. Those are \emph{basic} residual blocks used in \resnet{-18} and \resnet{-34}; for \resnet{-50} and more layers, the \emph{bottleneck} residual blocks are used, which are similar to the illustrations here except the \textsf{residual body} is now $\mathsf{C\rightarrow N \rightarrow R\rightarrow C\rightarrow N\rightarrow R\rightarrow C}$ with a $4\times$ reduction of the convolution channels in the middle for a ``bottlenecked'' residual.}
  \label{fig:resblk-illustration}
\end{figure}

We mainly study three types of neural network architectures:
\begin{itemize}

  \item \fcn{s}: the \fcn{s} consist of fully connected layers
    with equal output dimension and ReLU activations (except for the last layer,
    where the output dimension equals the number of classes and no ReLU is
    applied). For example, \fcn{ $3\times 256$} has three layers of fully
    connected layers with the output dimension 256, and an extra final (fully
    connected) classifier layer with the output dimension 10 (for \cifar and \mnist).

  \item \vgg{s}: widely used network architectures from \citet{simonyan2014very},
    consist of multiple convolutional layers, followed by multiple fully connected
    layers and the final linear classifier layer.

  \item \resnet{s}: the results from our analysis are similar for \resnet{s} V1
    \citep{he2016deep} and V2 \citep{he2016identity}. We report our results
    with \resnet{s} V2 due to the slightly better performance.
    For large image sizes from \imagenet{}, the \texttt{stage0} contains
    a $7\times 7$ convolution and a $3\times 3$ max pooling (both with stride
    2) to reduce the spatial dimension (from 224 to 56). On smaller image sizes
    like \cifar{}, we use a $3\times 3$ convolution with stride 1 here to avoid
    reducing the spatial dimension. Fig.~\ref{fig:resblk-illustration}
    illustrates the two types of residual blocks that are used inside (with an
    \emph{identity} skip connection) and between (with a \emph{downsample} skip connection)
    stages.

    In the experiments on adversarial robustness in
    Sec.~\ref{sec:other-robustness}, we use a slightly modified variant by
    explicitly having a downsample layer between stages, so that all the
    residual blocks are with \emph{identity} skip connections.

    The \resnet{s} used in the main text are \emph{without} batch normalization.
    Please see Appendix~\ref{app:batchnorm} for details and full comparison of the
    architectures with and without batch normalization.
\end{itemize}

During training, \cifar{} images are padded with 4 pixels of zeros on all sides, then randomly flipped (horizontally) and cropped. \imagenet{} images are randomly cropped during training and center-cropped during testing. Global mean and standard deviation are computed on all the training pixels and applied to normalize the inputs on each dataset.

\section{Further Details on Joint Robustness}
\label{app:joint-robustness}

\begin{figure}
  \begin{subfigure}{.32\linewidth}
    \includegraphics[width=\linewidth]{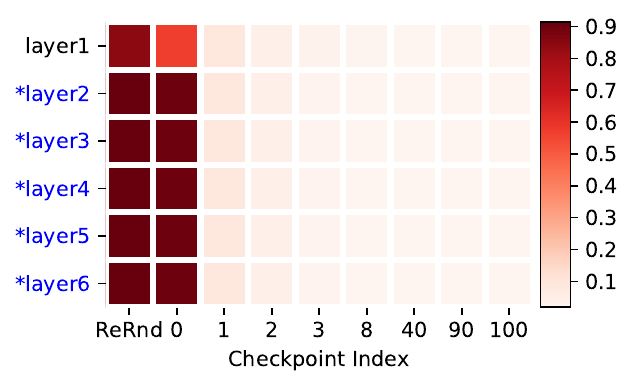}
    \caption{\scriptsize\textrm{layer2}$\sim$\texttt{6}}
  \end{subfigure}\hfill
  \begin{subfigure}{.32\linewidth}
    \includegraphics[width=\linewidth]{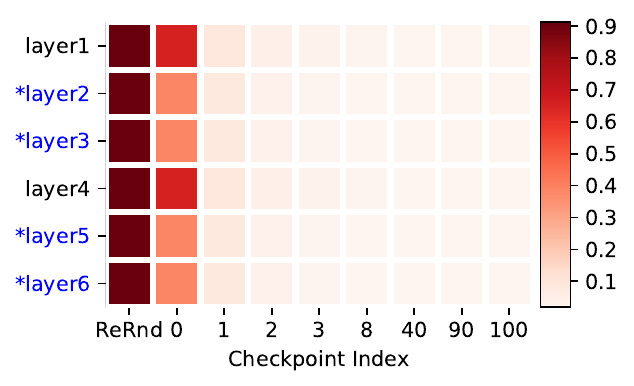}
    \caption{\scriptsize\textrm{layer2,3,5,6}}
  \end{subfigure}\hfill
  \begin{subfigure}{.32\linewidth}
    \includegraphics[width=\linewidth]{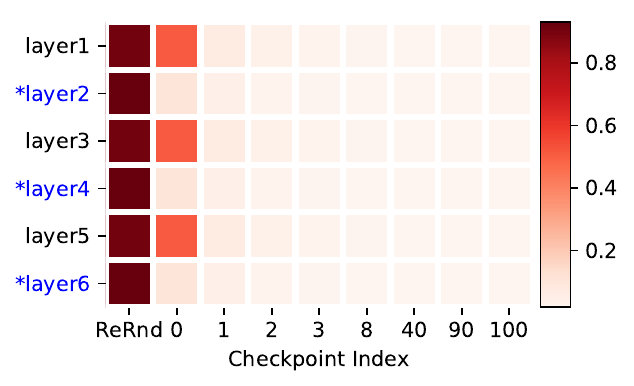}
    \caption{\scriptsize{\textrm{layer2,4,6}}}
  \end{subfigure}
  \caption{\textbf{Joint robustness analysis of \fcn{ $5\times 256$} on MNIST.}
  The heatmap layout is the same as in Fig.~\ref{fig:mnist-mlp-3x256}, but the
  layers are divided into two groups (indicated by the \texttt{*} mark on the
  \textcolor{blue}{blue colored} layer names in each figure) and
  re-randomization and re-initialization are applied to all the layers in each
  group \emph{jointly}. As a result, layers belonging to the same group have
  identical rows in the heatmap, but we still show all the layers to make the
  figures easier to compare with the previous individual layer
  robustness results. The subfigures show the results from three different
  grouping schemes.}
  \label{fig:mnist-bulk}
\end{figure}

\begin{figure}
  \begin{subfigure}{.38\linewidth}
    \includegraphics[width=\linewidth]{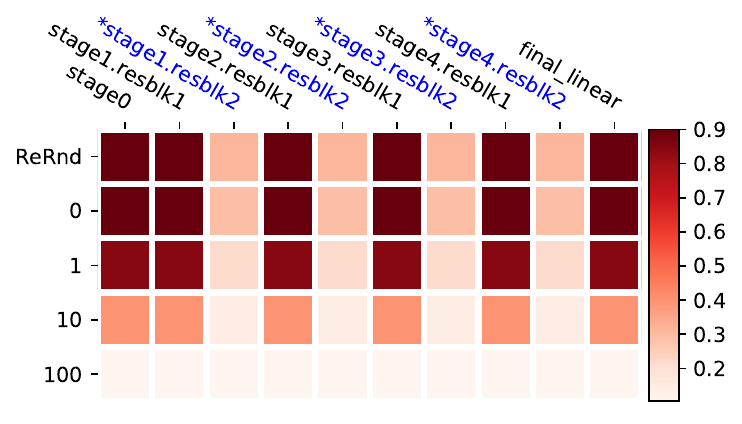}
    \caption{\scriptsize{\resnet{-18}: \textrm{resblk2}}}
  \end{subfigure}
  \begin{subfigure}{.61\linewidth}
    \includegraphics[width=\linewidth]{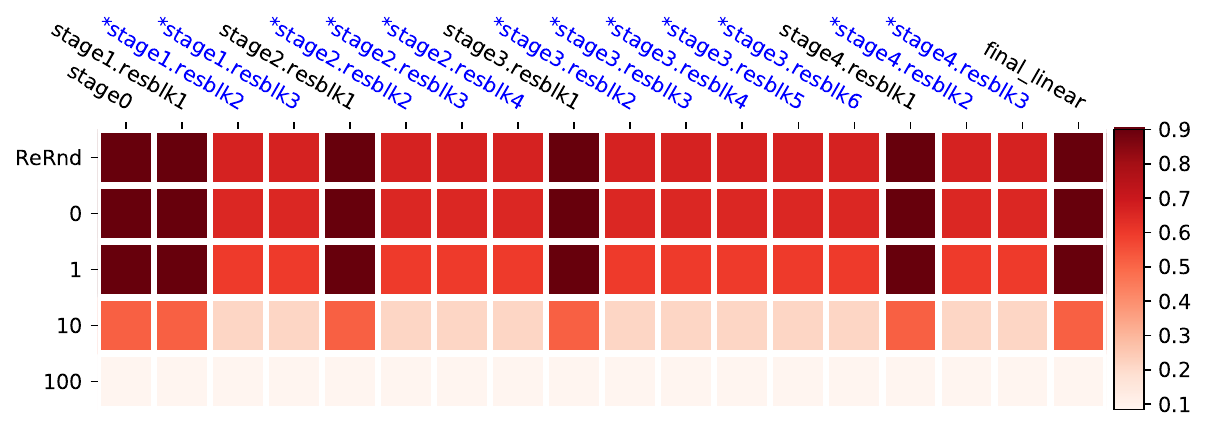}
    \caption{\scriptsize\resnet{-50}: \textrm{resblk2, 3, }\ldots}
  \end{subfigure}
  \begin{subfigure}{\linewidth}
    \includegraphics[width=\linewidth]{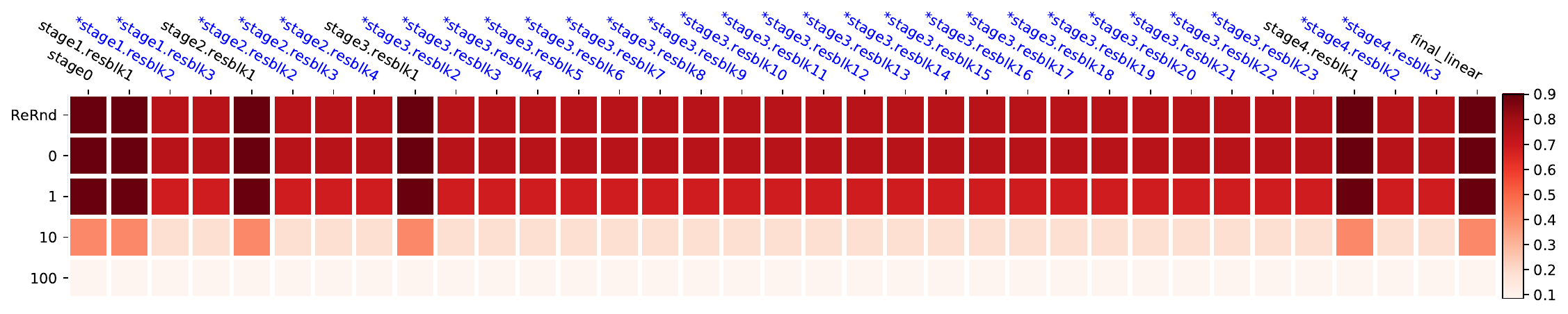}
    \caption{\scriptsize\resnet{-101}: \textrm{resblk2, 3, }\ldots}
  \end{subfigure}
  \begin{subfigure}{\linewidth}
    \includegraphics[width=\linewidth]{figs/bulk-cifar-with-fullmodel/resnet-152-resblk1-tt-error}
    \caption{\scriptsize\resnet{-152}: \textrm{resblk2, 3, }\ldots}
  \end{subfigure}
  \caption{\textbf{Joint robustness analysis of \resnet{s} on \cifar}, based on the
  scheme that groups all but the first residual blocks in all the stages. Grouping is
	indicated by the \texttt{*} on the (\textcolor{blue}{blue colored}) layer names.}
  \label{fig:bulk-resnet}
\end{figure}

\begin{figure}
  \begin{subfigure}{.49\linewidth}
    \includegraphics[width=\linewidth]{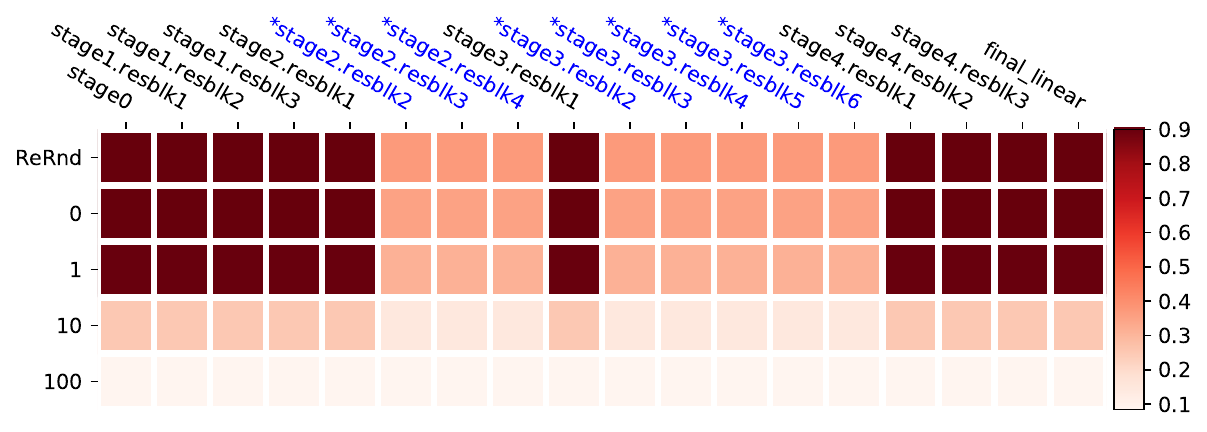}
    \caption{\scriptsize\resnet{-50}: \textrm{resblk2, 3 }\ldots of \textrm{stage2, 3}}
  \end{subfigure}
  \begin{subfigure}{.49\linewidth}
    \includegraphics[width=\linewidth]{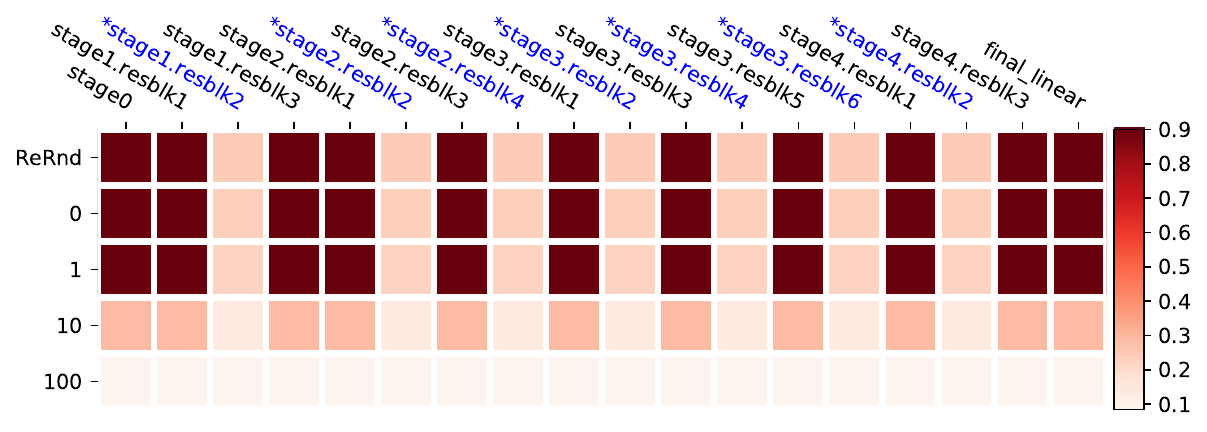}
    \caption{\scriptsize\resnet{-50}: every second \textrm{resblk}}
  \end{subfigure}
  \begin{subfigure}{\linewidth}
    \includegraphics[width=\linewidth]{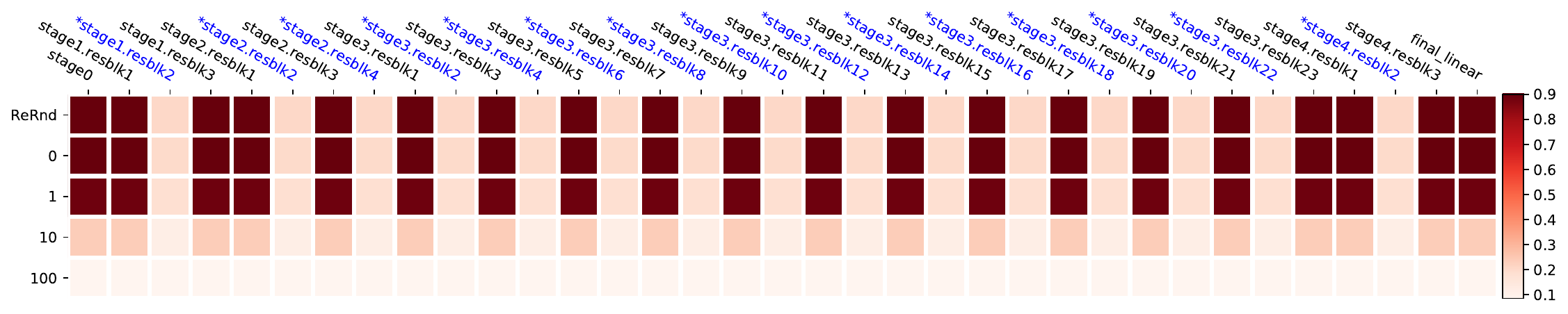}
    \caption{\scriptsize\resnet{-101}: every second \textrm{resblk}}
  \end{subfigure}
  \begin{subfigure}{\linewidth}
    \includegraphics[width=\linewidth]{figs/bulk-cifar-with-fullmodel/resnet-152-every2-tt-error}
    \caption{\scriptsize\resnet{-152}: every second \textrm{resblk}}
  \end{subfigure}
  \caption{\textbf{Joint robustness analysis of \resnet{s} on \cifar}, with alternative
  grouping schemes. Grouping is indicated by the \texttt{*} on the (\textcolor{blue}{blue colored}) layer names.}
  \label{fig:bulk-resnet-alt-scheme}
\end{figure}

In this appendix, we provide results on joint robustness analysis that were not
included in the main text due to space limit. From
Sec.~\ref{sec:mnist-mlp}, we see that on MNIST, for wide enough \fcn{s}, all
the layers above \texttt{layer1} are robust to re-initialization. So we divide
the layers into two groups: \{\texttt{layer1}\} and \{\texttt{layer2},
\texttt{layer3}, \ldots\}, and perform the robustness study on the two
groups. The results for \fcn{~$5\times 256$} are shown in
Fig.~\ref{fig:mnist-bulk}(a). For clarity and ease of comparison, the figure
still spells out all the layers individually, but the values from
\texttt{layer2} to \texttt{layer6} are simply repeated rows. The values show
that the upper-layer-group is clearly \emph{not} jointly robust to
re-initialization (to checkpoint-0).

We also try some alternative grouping schemes: Fig.~\ref{fig:mnist-bulk}(b)
show the results when we group two in every three layers, which has slightly
improved joint robustness. In Fig.~\ref{fig:mnist-bulk}(c), the grouping
scheme that includes every other layer shows that with a clever grouping
scheme, about half of the layers could be \emph{jointly} robust.

Results on \resnet{s} are similar. Fig.~\ref{fig:bulk-resnet} shows the joint
robustness analysis on \resnet{s} trained on \cifar. The grouping is based on
the individual layer robustness results from Fig.~\ref{fig:cifar10-resnet}: all
the residual blocks in \texttt{stage1} to \texttt{stage4} are bundled and
analyzed jointly. The results are similar to the \fcn{s}: \resnet{-18} is
relatively robust, but deeper \resnet{s} are \emph{not} jointly robust under this
grouping. Two alternative grouping schemes are shown in
Fig.~\ref{fig:bulk-resnet-alt-scheme}. By including only layers from
\texttt{stage1} and \texttt{stage4}, slightly improved robustness could be
obtained on \resnet{-50}. The scheme that groups every other residual block
shows further improvements.

In summary, the individually robust layers are generally not jointly robust.
But with some clever way of picking out a subset of the layers, joint
robustness could still be achieved for up to half of the layers. In principle,
one can enumerate all possible grouping schemes to find the best with a
trade-off of the robustness and number of layers included.

\section{Batch Normalization and Weight Decay}
\label{app:batchnorm}

The primary goal of this paper is to study the (co-)evolution of the representations at each layer during training and the robustness of this representation with respect to the rest of the network. We try to minimize the factors that explicitly encourage changing of the network weights or representations in the analysis. In particular, unless otherwise specified, weight decay and batch normalization were \emph{not} used. This leads to some performance drop in the trained models. Especially for deep residual networks on \imagenet: even though we could successfully train a residual network with 100+ layers without batch normalization, the final generalization performance could be quite worse than the state-of-the-art. Therefore, in this section, we include experiments with networks trained \emph{with} weight decay and batch normalization for comparison.

\begin{table}\centering
  \caption{\textbf{Test performance (classification error rates \%) of various models studied in this paper.} The table shows how much of the final performance is affected by training with or without weight decay (+wd) and batch normalization (+bn).}
  \label{tab:wd-bn-comparison}
  \begin{tabular}{rlcccc}\toprule
  & Architecture & N/A & +wd & +bn & +wd+bn \\\midrule
  \multirow{9}{*}{\rotatebox[origin=c]{90}{\cifar}}
  &\resnet{-18} & 10.4 & 7.5 & 6.9 & 5.5 \\
  &\resnet{-34} & 10.2 & 6.9 & 6.6 & 5.1 \\
  &\resnet{-50} & 8.4 & 9.9 & 7.6 & 5.0 \\
  &\resnet{-101} & 8.5 & 9.8 & 6.9 & 5.3 \\
  &\resnet{-152} & 8.5 & 9.7 & 7.3 & 4.7 \\
  &\vgg{-11} & 11.8 & 10.7 & 9.4 & 8.2 \\
  &\vgg{-13} & 10.3 & 8.8 & 8.4 & 6.7 \\
  &\vgg{-16} & 11.0 & 11.4 & 8.5 & 6.7 \\
  &\vgg{-19} & 12.1 & & 8.6 & 6.9 \\
  \midrule
  \multirow{5}{*}{\rotatebox[origin=c]{90}{\imagenet}}
  &\resnet{-18} & 41.1 & 33.1 & 33.5 & 31.5 \\
  &\resnet{-34} & 39.9 & 30.6 & 30.1 & 27.2 \\
  &\resnet{-50} & 34.8 & 31.8 & 28.2 & 25.0 \\
  &\resnet{-101} & 32.9 & 29.9 & 26.9 & 22.9 \\
  &\resnet{-152} & 31.9 & 29.1 & 27.6 & 22.6 \\
  \bottomrule
  \end{tabular}
\end{table}

Table~\ref{tab:wd-bn-comparison} shows the final test error
rates of models trained with or without weight decay and batch normalization.
Note the original \vgg{} models do not use batch normalization
\citep{simonyan2014very}, we list +bn variants here for comparison, by applying
batch normalization to the output of each convolutional layer. On \cifar{}, the
performance gap varies from 3\% to 5\%, but on \imagenet{}, the gap
could be as large as 10\%.

\begin{figure}
    \begin{subfigure}{.49\linewidth}
      \includegraphics[width=\linewidth]{figs/vgg-hm-with-fullmodel/vgg16-tt-error}
      \caption{\vgg{-16}}
    \end{subfigure}
    \begin{subfigure}{.49\linewidth}
      \includegraphics[width=\linewidth]{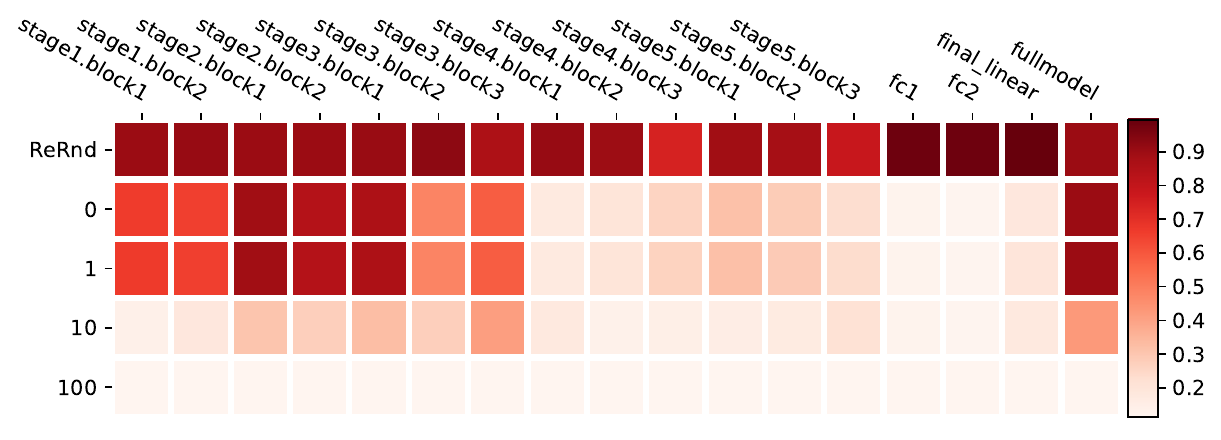}
      \caption{\vgg{-16} +wd}
    \end{subfigure}
    \caption{\textbf{Layer robustness analysis with \vgg{16} on \cifar.} The subfigures show how training with weight decay (+wd)
    affects the layer robustness patterns.}
    \label{fig:vgg-variants}
\end{figure}

\begin{figure}
    \begin{subfigure}{.49\linewidth}
      \includegraphics[width=\linewidth]{figs/resnet-hm-with-fullmodel/cifar-small-50-tt-error}
      \caption{\resnet{-50}}
    \end{subfigure}
    \begin{subfigure}{.49\linewidth}
      \includegraphics[width=\linewidth]{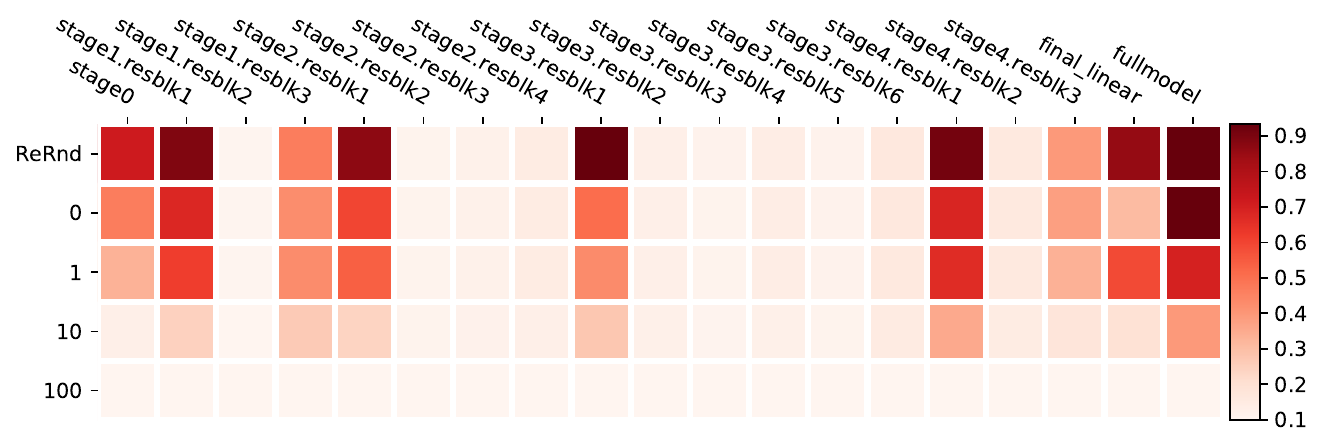}
      \caption{\resnet{-50} +wd}
    \end{subfigure}
    \begin{subfigure}{.49\linewidth}
      \includegraphics[width=\linewidth]{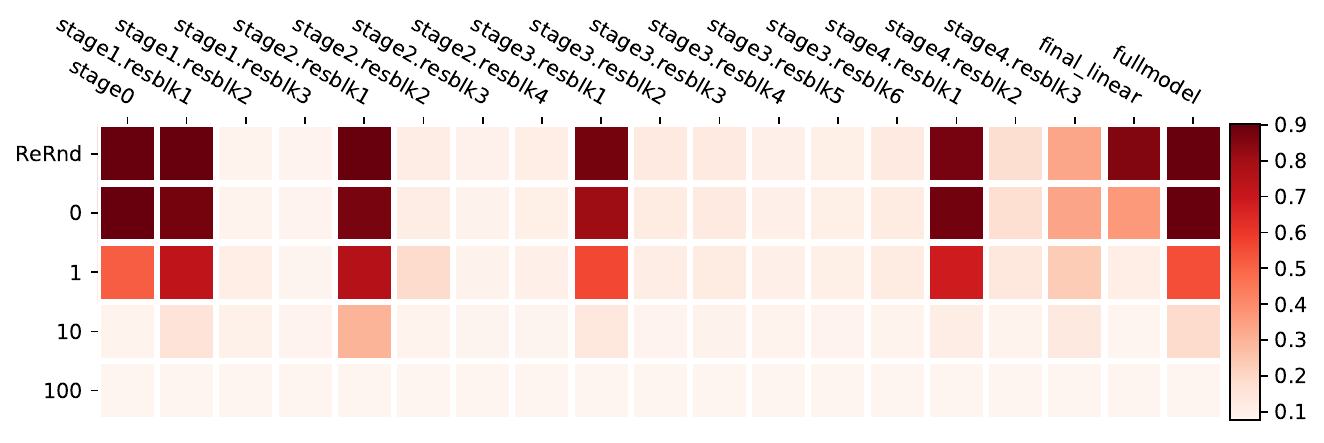}
      \caption{\resnet{-50} +bn}
    \end{subfigure}
    \begin{subfigure}{.49\linewidth}
      \includegraphics[width=\linewidth]{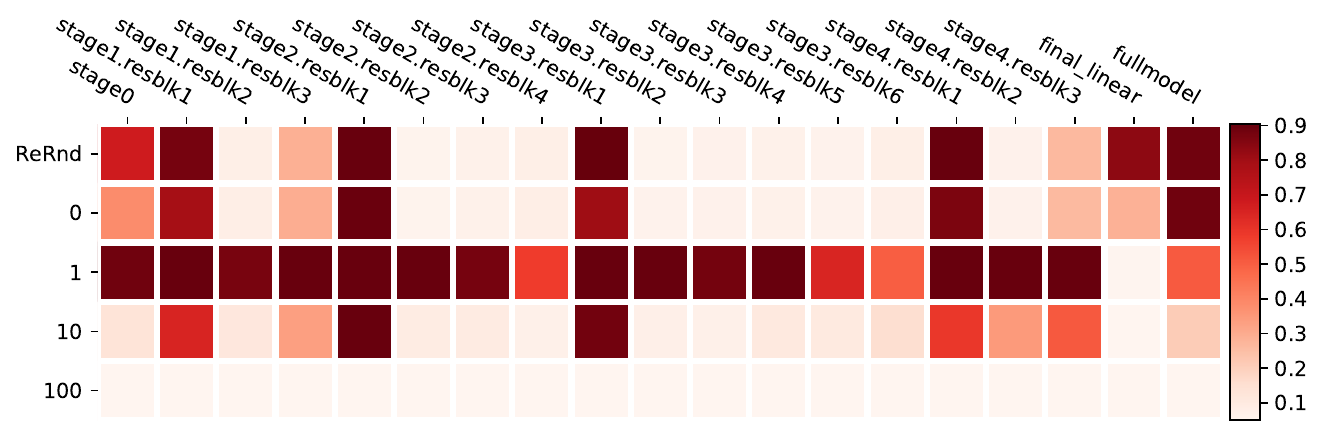}
      \caption{\resnet{-50} +wd +bn}
    \end{subfigure}
    \caption{\textbf{Layer robustness analysis with \resnet{-50} on \cifar.} The subfigures show how training with weight decay (+wd) and batch normalization (+bn) affects the layer robustness patterns.}
    \label{fig:cifar-resnet-variants}
\end{figure}

\begin{figure}
    \begin{subfigure}{.49\linewidth}
      \includegraphics[width=\linewidth]{figs/resnet-hm-with-fullmodel/imagenet-50-tt-error}
      \caption{\resnet{-50}}
    \end{subfigure}
    \begin{subfigure}{.49\linewidth}
      \includegraphics[width=\linewidth]{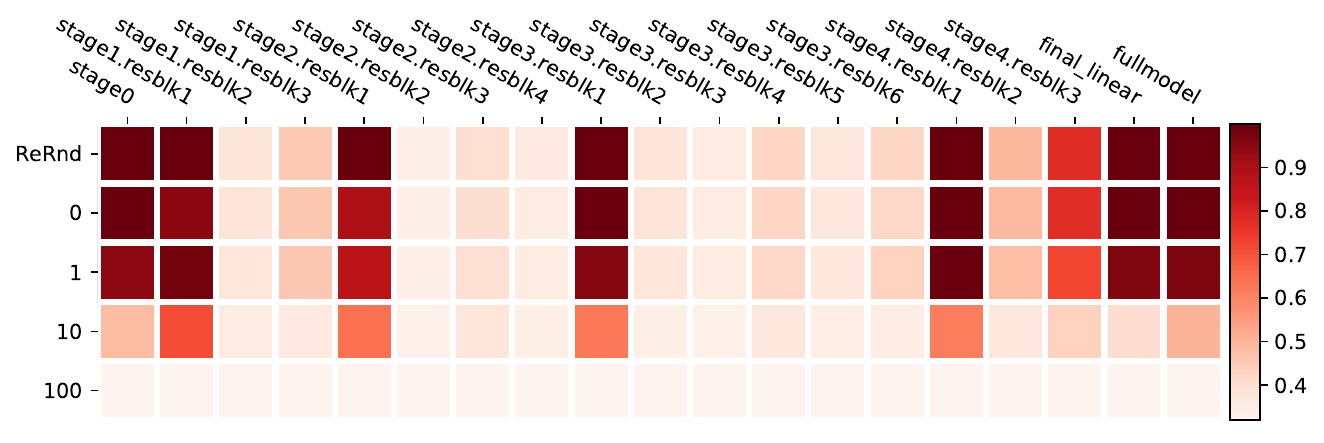}
      \caption{\resnet{-50} +wd}
    \end{subfigure}
    \begin{subfigure}{.49\linewidth}
      \includegraphics[width=\linewidth]{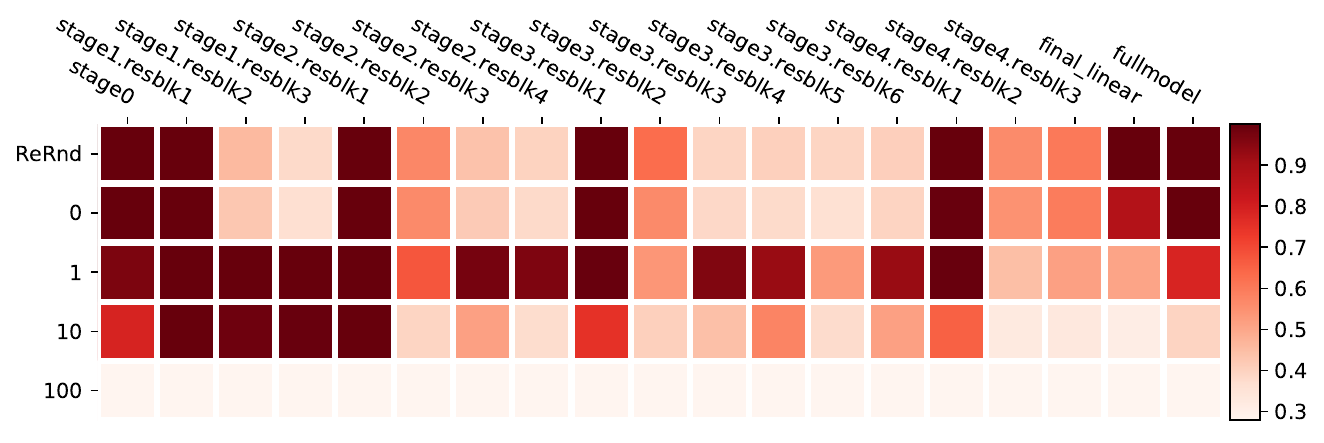}
      \caption{\resnet{-50} +bn}
    \end{subfigure}
    \begin{subfigure}{.49\linewidth}
      \includegraphics[width=\linewidth]{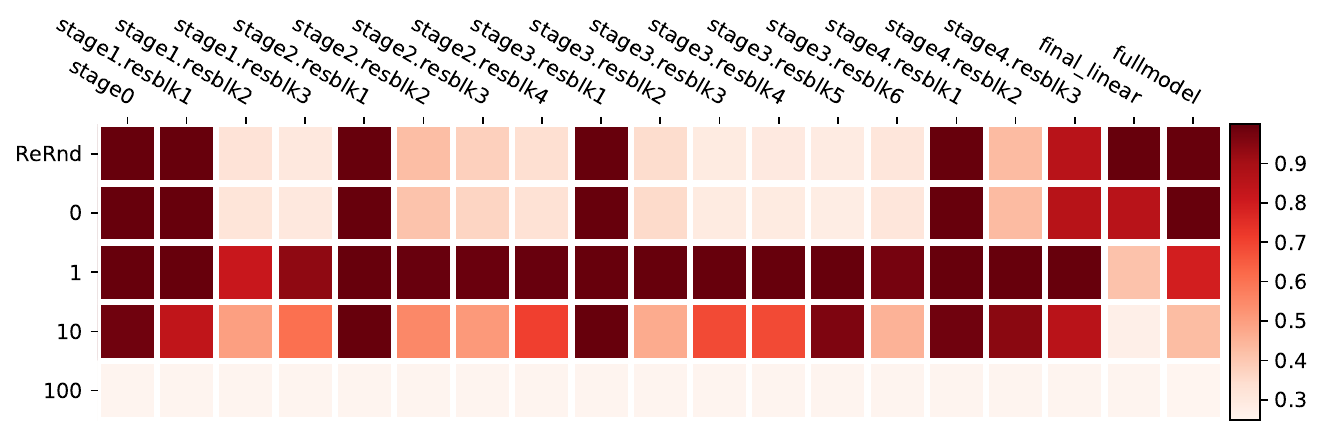}
      \caption{\resnet{-50} +wd +bn}
    \end{subfigure}
    \caption{\textbf{Layer robustness analysis with \resnet{-50} on \imagenet.} The subfigures show how training with weight decay (+wd) and batch normalization (+bn) affects the layer robustness patterns.}
    \label{fig:imagenet-resnet-variants}
\end{figure}

Fig.~\ref{fig:vgg-variants} shows how different training
configurations affect the layer robustness analysis patterns on \vgg{-16}
networks.
Fig.~\ref{fig:cifar-resnet-variants} and Fig.~\ref{fig:imagenet-resnet-variants} show similar comparisons for \resnet{-50} on \cifar{} and \imagenet{}, respectively. We found that the layer robustness patterns are still quite pronounced under various training conditions. In Fig.~\ref{fig:cifar-resnet-variants}(d) and Fig.~\ref{fig:imagenet-resnet-variants}(c,d), we found that re-initialing with checkpoint-1 is less robust than with checkpoint-0 for many layers.
It might be that during early stages, some aggressive learning is causing changes in the parameters or statistics with large magnitudes, but later on when most of the training samples are classified correctly, the network gradually re-balances the layers to a more robust state. Fig.~\ref{fig:app-cifar-resnet}(d-f) in the next section show supportive evidence that, in this case the distance of the parameters between checkpoint-0 and checkpoint-1 is larger than between checkpoint-0 and the final checkpoint-T. However, on \imagenet{} this correlation is no longer clear as in Fig.~\ref{fig:app-imagenet-resnet}(d-f). See also the discussions in the next section.

\section{Robustness and Distances}

\begin{figure*}
  \begin{subfigure}{.32\linewidth}
    \includegraphics[width=\linewidth]{figs/vgg-hm-with-fullmodel/vgg16-tt-error}
    \caption{Test error}
  \end{subfigure}
  \begin{subfigure}{.32\linewidth}
    \includegraphics[width=\linewidth]{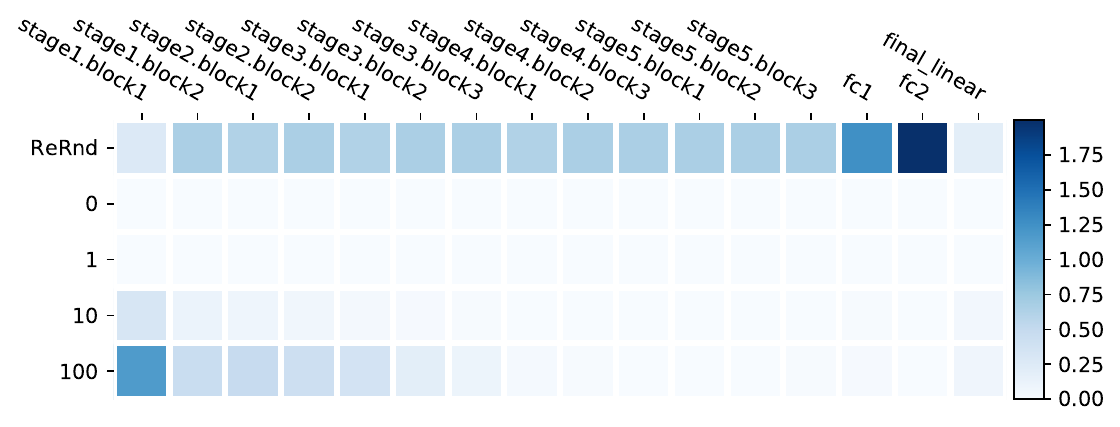}
    \caption{$\|\theta_d^{\tau}-\theta_d^{0}\|$}
  \end{subfigure}
  \begin{subfigure}{.32\linewidth}
    \includegraphics[width=\linewidth]{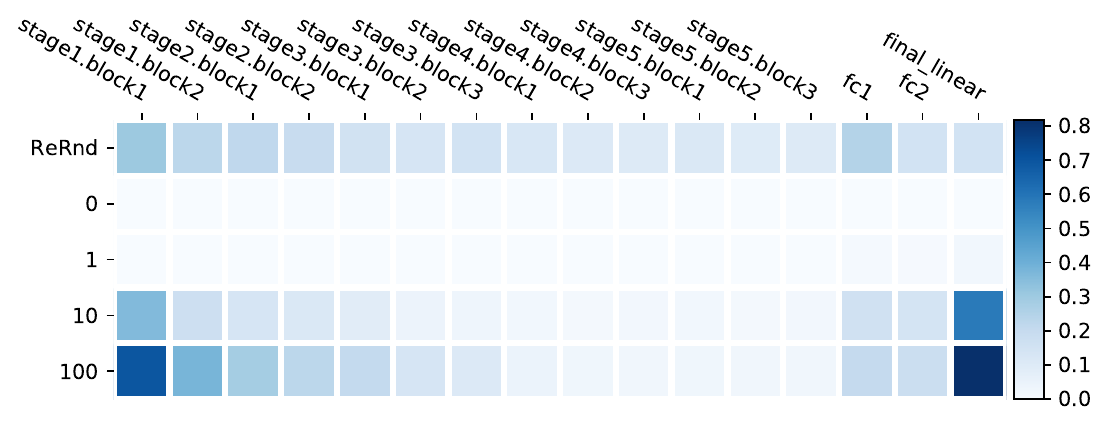}
    \caption{$\|\theta_d^{\tau}-\theta_d^{0}\|_\infty$}
  \end{subfigure}
  \caption{\textbf{Layer robustness of \vgg{-16} on \cifar.} (a) shows the robustness analysis measured by the test error rate. (b) shows the normalized $\ell_2$ distance of the parameters at each layer to the version realized during the re-randomization and re-initialization analysis. (c) is the same as (b), except with the $\ell_\infty$ distance.}
  \label{fig:app-vgg}
\end{figure*}

\begin{figure*}
  \begin{subfigure}{.32\linewidth}
    \includegraphics[width=\linewidth]{figs/resnet-hm-with-fullmodel/cifar-small-50-tt-error}
    \caption{Test error (-wd-bn)}
  \end{subfigure}
  \begin{subfigure}{.32\linewidth}
    \includegraphics[width=\linewidth]{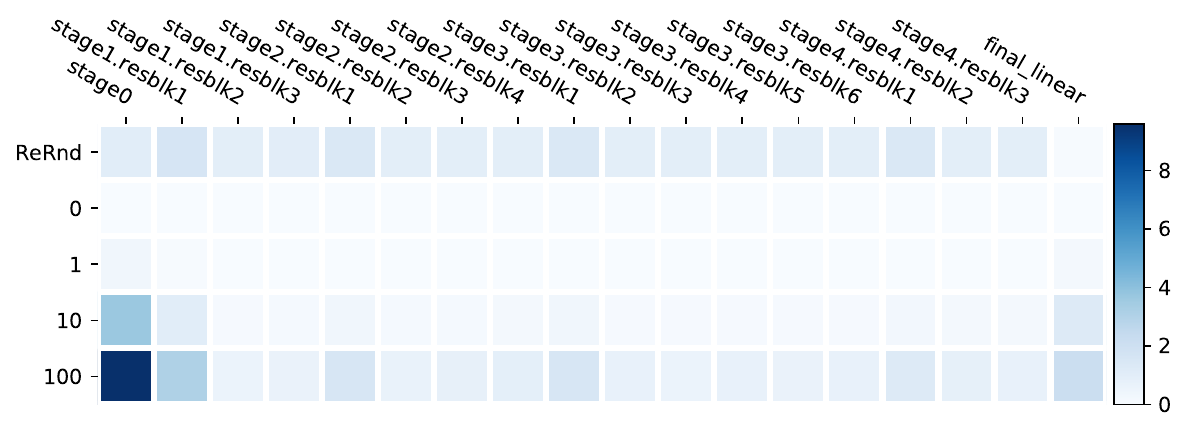}
    \caption{$\|\theta_d^{\tau}-\theta_d^{0}\|$}
  \end{subfigure}
  \begin{subfigure}{.32\linewidth}
    \includegraphics[width=\linewidth]{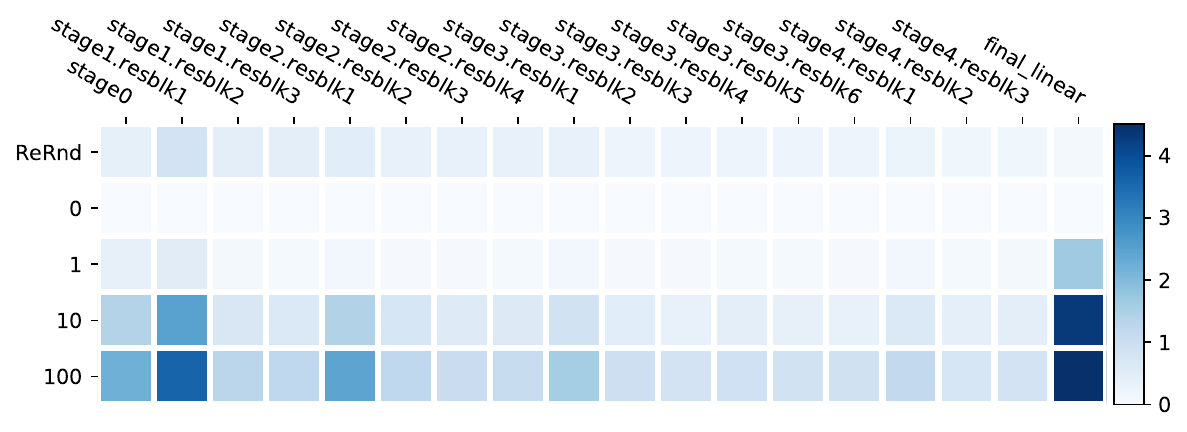}
    \caption{$\|\theta_d^{\tau}-\theta_d^{0}\|_\infty$}
  \end{subfigure}\\
  \begin{subfigure}{.32\linewidth}
    \includegraphics[width=\linewidth]{figs/resnet-hm-with-fullmodel/cifar-small-50-wd-bn-AS-tt-error}
    \caption{Test error (+wd+bn)}
  \end{subfigure}
  \begin{subfigure}{.32\linewidth}
    \includegraphics[width=\linewidth]{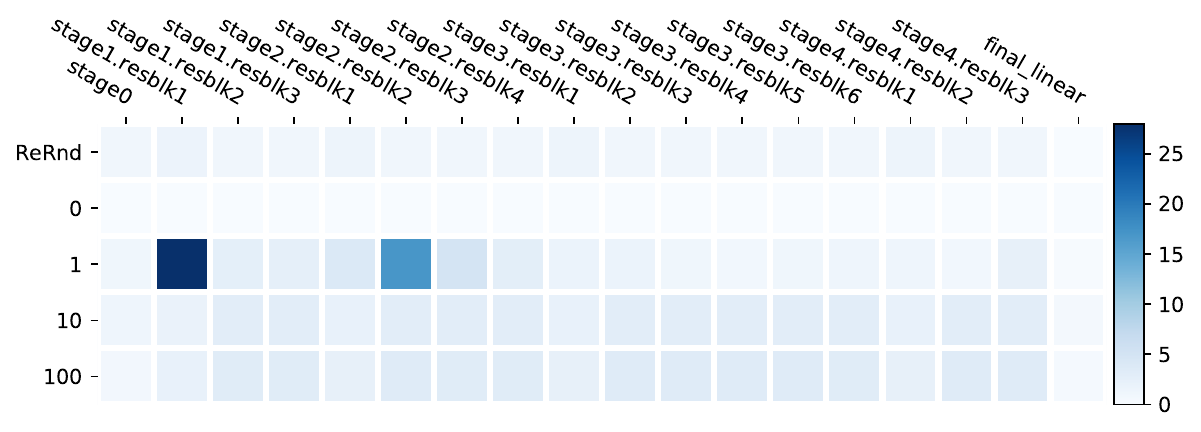}
    \caption{$\|\theta_d^{\tau}-\theta_d^{0}\|$}
  \end{subfigure}
  \begin{subfigure}{.32\linewidth}
    \includegraphics[width=\linewidth]{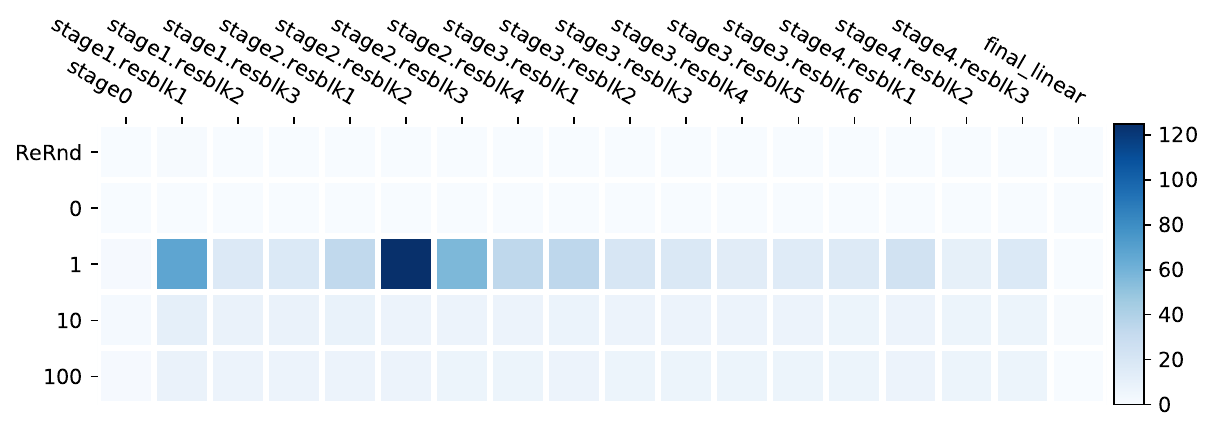}
    \caption{$\|\theta_d^{\tau}-\theta_d^{0}\|_\infty$}
  \end{subfigure}
  \caption{\textbf{Layer robustness for \resnet{-50} on \cifar.} Layouts are the same as
  in Fig.~\ref{fig:app-vgg}. The first row (a-c) is for \resnet{-50} trained
  without weight decay and batch normalization. The second row (d-f) is with
  weight decay and batch normalization.}
  \label{fig:app-cifar-resnet}
\end{figure*}

\begin{figure*}
  \begin{subfigure}{.32\linewidth}
    \includegraphics[width=\linewidth]{figs/resnet-hm-with-fullmodel/imagenet-50-tt-error}
    \caption{Test error (-wd-bn)}
  \end{subfigure}
  \begin{subfigure}{.32\linewidth}
    \includegraphics[width=\linewidth]{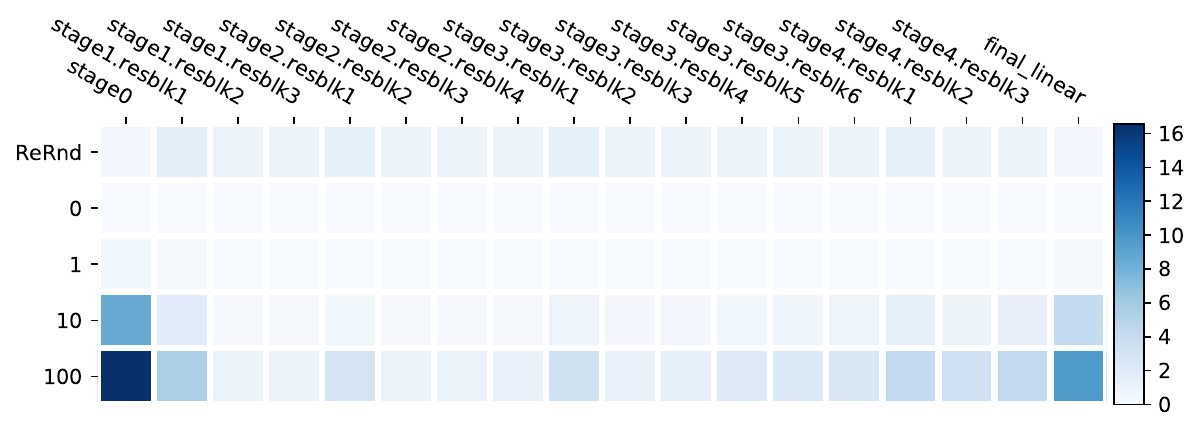}
    \caption{$\|\theta_d^{\tau}-\theta_d^{0}\|$}
  \end{subfigure}
  \begin{subfigure}{.32\linewidth}
    \includegraphics[width=\linewidth]{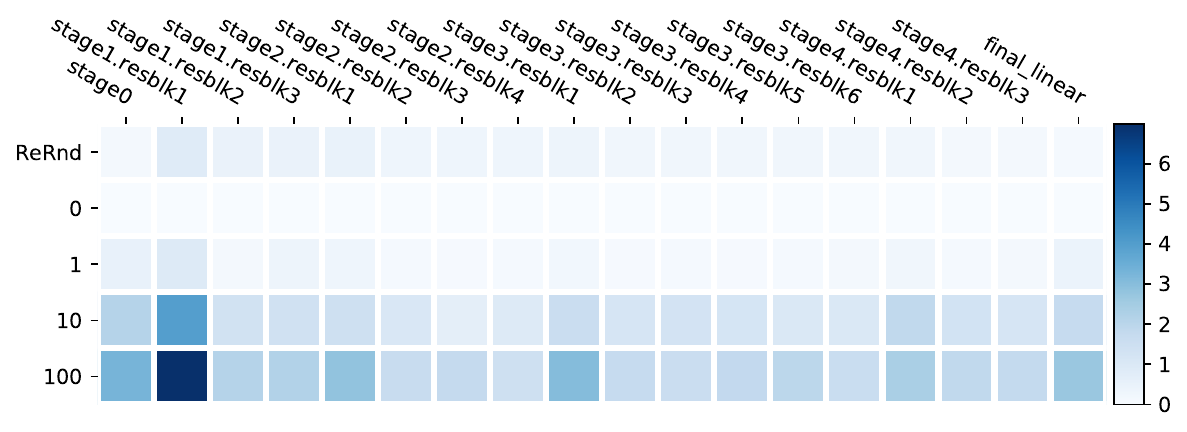}
    \caption{$\|\theta_d^{\tau}-\theta_d^{0}\|_\infty$}
  \end{subfigure}\\
  \begin{subfigure}{.32\linewidth}
    \includegraphics[width=\linewidth]{figs/resnet-hm-with-fullmodel/imagenet-50-wd-bn-AS-tt-error}
    \caption{Test error (+wd+bn)}
  \end{subfigure}
  \begin{subfigure}{.32\linewidth}
    \includegraphics[width=\linewidth]{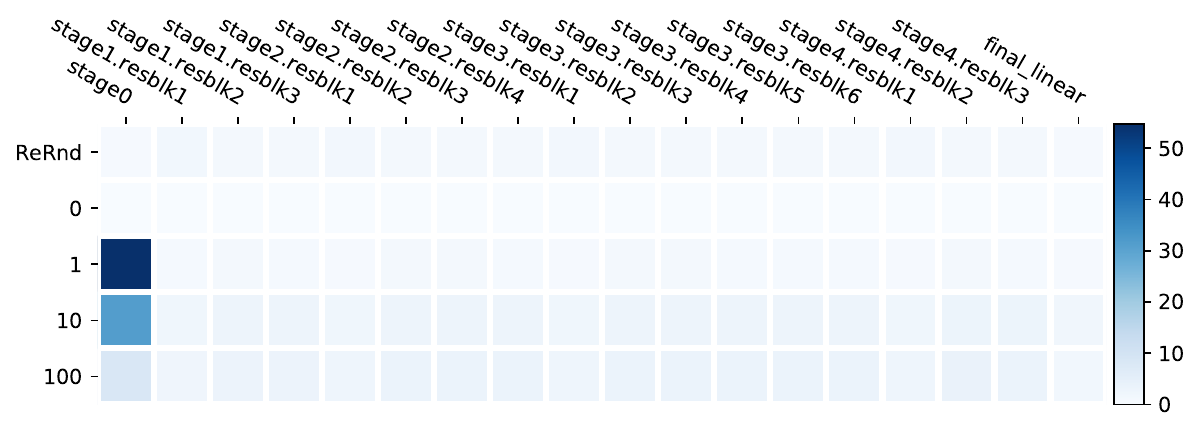}
    \caption{$\|\theta_d^{\tau}-\theta_d^{0}\|$}
  \end{subfigure}
  \begin{subfigure}{.32\linewidth}
    \includegraphics[width=\linewidth]{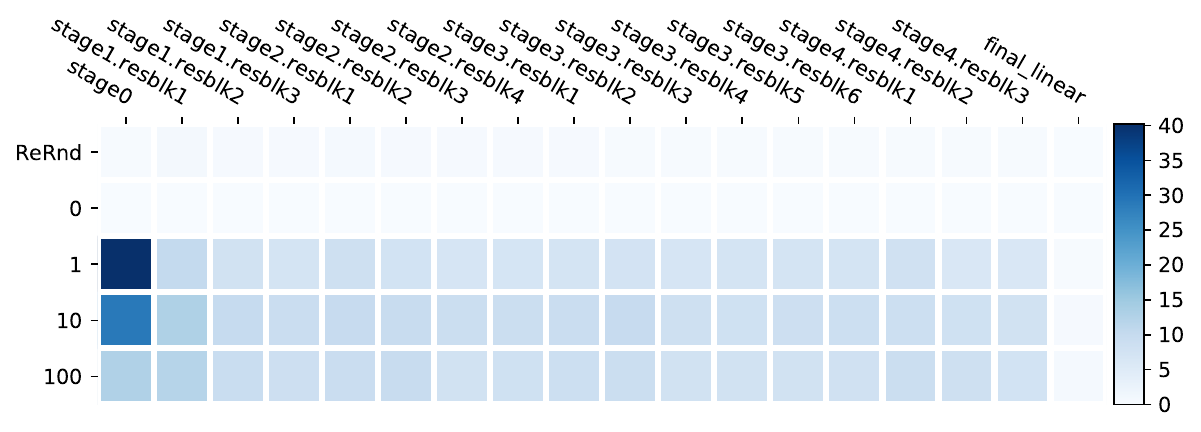}
    \caption{$\|\theta_d^{\tau}-\theta_d^{0}\|_\infty$}
  \end{subfigure}
  \caption{\textbf{Layer robustness  of \resnet{-50} on \imagenet.} Layouts are the same as in Fig.~\ref{fig:app-vgg}. The first row (a-c) is for \resnet{-50} trained without weight decay and batch normalization. The second row (d-f) is with weight decay and batch normalization.}
  \label{fig:app-imagenet-resnet}
\end{figure*}

In Fig.~\ref{fig:mnist-mlp-3x256} from Sec.~\ref{sec:mnist-mlp}, we
compared the layer robustness patterns to the layer-wise distances of the
parameters to the values at initialization (checkpoint-0). We found that for
\fcn{s} on \mnist, there is no obvious correlation between the ``amount of
parameter updates received'' at each layer and its robustness to
re-initialization for the two distances (the normalized $2$ and
$\infty$ norms) we measured. In this appendix, we list results on
other models and datasets studied in this paper for comparison.

Fig.~\ref{fig:app-vgg} shows the layer robustness plot along with the layer-wise distance plots for \vgg{-16} trained on \cifar. We found that the $\ell_\infty$ distance of the top layers are large, but the model is robust when we re-initialize those layers. However, the normalized $\ell_2$ distance seem to be correlated with the layer robustness patterns: the upper layers that are more robust moved smaller distances from their initialized values during training.

Similar plots for \resnet{-50} on \cifar{} and \imagenet{} are shown in Fig.~\ref{fig:app-cifar-resnet} and Fig.~\ref{fig:app-imagenet-resnet}, respectively. In each of the figures, we also show extra results for models trained with weight decay and batch normalization. For the case without weight decay and batch normalization, we can see a weak correlation: the layers that are \critical have slightly larger distances to their random initialization values. For the case with weight decay and batch normalization, the situation is less clear. First of all, in Fig.~\ref{fig:app-cifar-resnet}(e-f), we see very large distances in a few layers at checkpoint-1. This provides a potential explanation to the mysterious pattern that re-initialization to checkpoint-1 is more sensitive than to checkpoint-0. Similar observations can be found in Fig.~\ref{fig:app-imagenet-resnet}(e-f) for \imagenet{}.

\section{Alternative Visualizations}

\begin{figure}[t]  \begin{subfigure}[t]{.32\linewidth}
    \includegraphics[width=\linewidth]{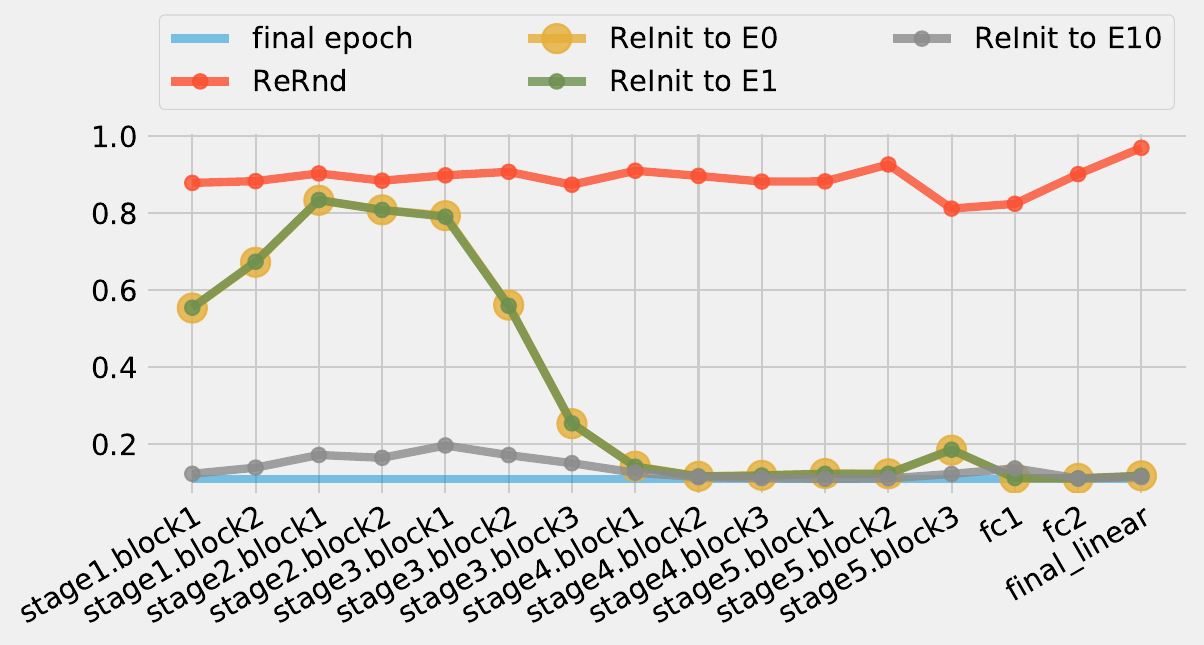}
    \caption{Test error}
  \end{subfigure}
  \begin{subfigure}[t]{.32\linewidth}
    \includegraphics[width=\linewidth]{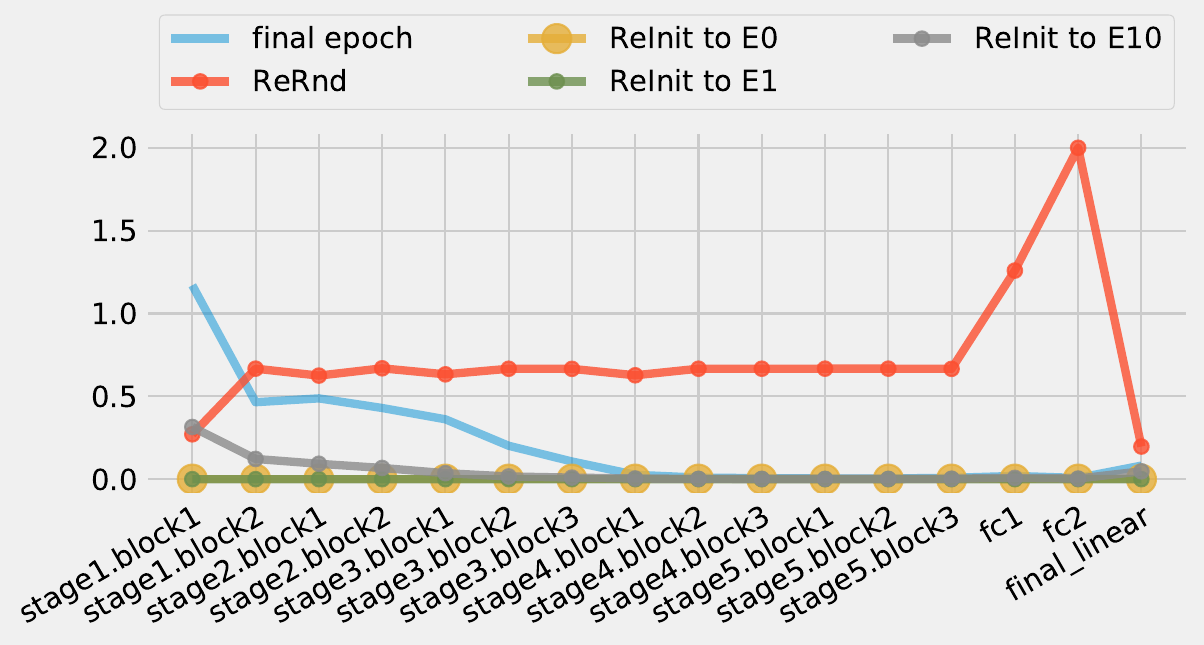}
    \caption{$\|\theta_d^{\tau}-\theta_d^{0}\|$}
  \end{subfigure}
  \begin{subfigure}[t]{.32\linewidth}
    \includegraphics[width=\linewidth]{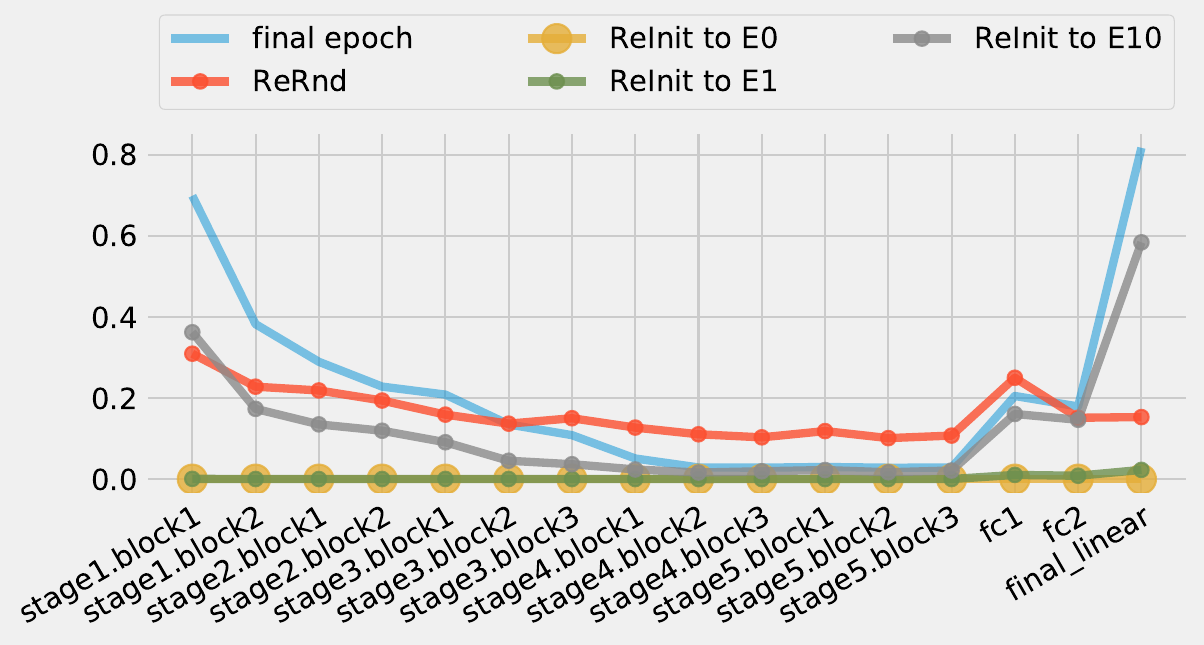}
    \caption{$\|\theta_d^{\tau}-\theta_d^{0}\|_\infty$}
  \end{subfigure}
  \caption{\textbf{Alternative visualization of layer robustness analysis for \vgg{-16} models on \cifar{}}. This shows the same results as Fig.~\ref{fig:app-vgg}, but shown as curves instead of heatmaps.}
  \label{fig:app-reset-curves-vgg16}
\end{figure}

\begin{figure}[t]  \begin{subfigure}[t]{.32\linewidth}
    \includegraphics[width=\linewidth]{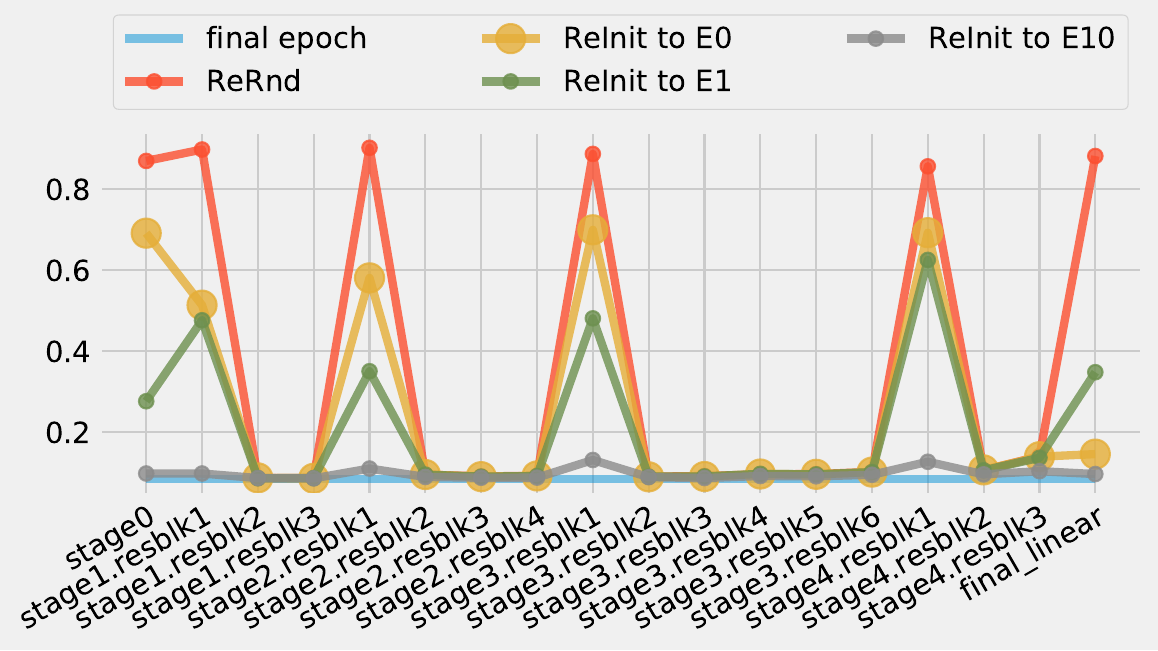}
    \caption{\cifar test error}
  \end{subfigure}
  \begin{subfigure}[t]{.32\linewidth}
    \includegraphics[width=\linewidth]{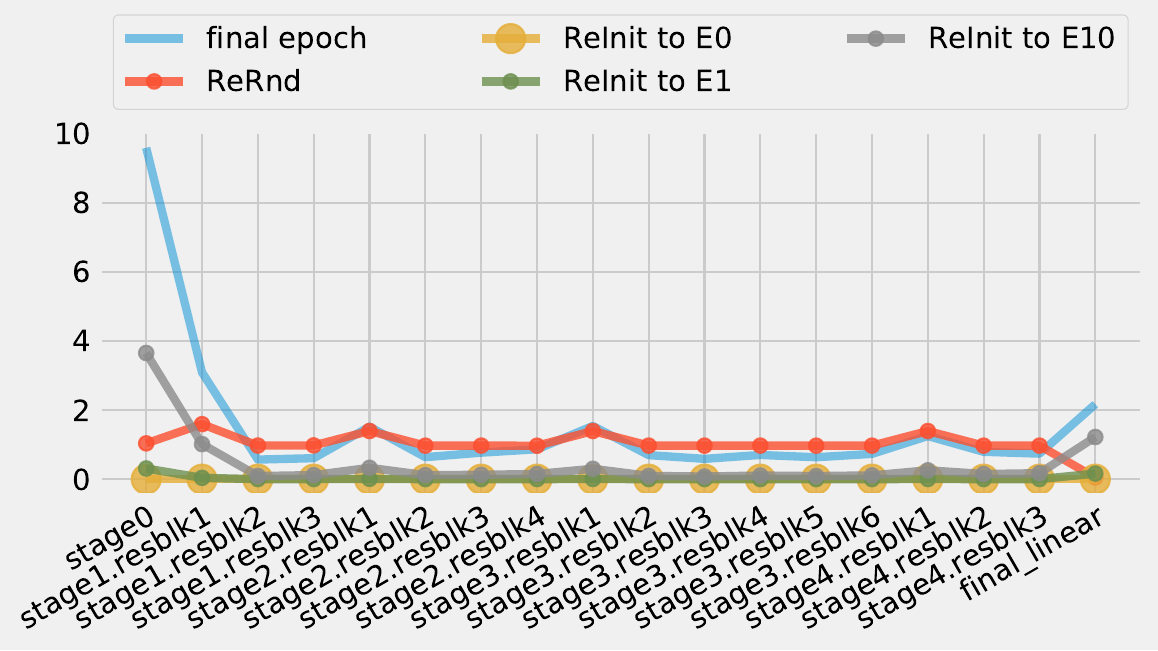}
    \caption{$\|\theta_d^{\tau}-\theta_d^{0}\|$}
  \end{subfigure}
  \begin{subfigure}[t]{.32\linewidth}
    \includegraphics[width=\linewidth]{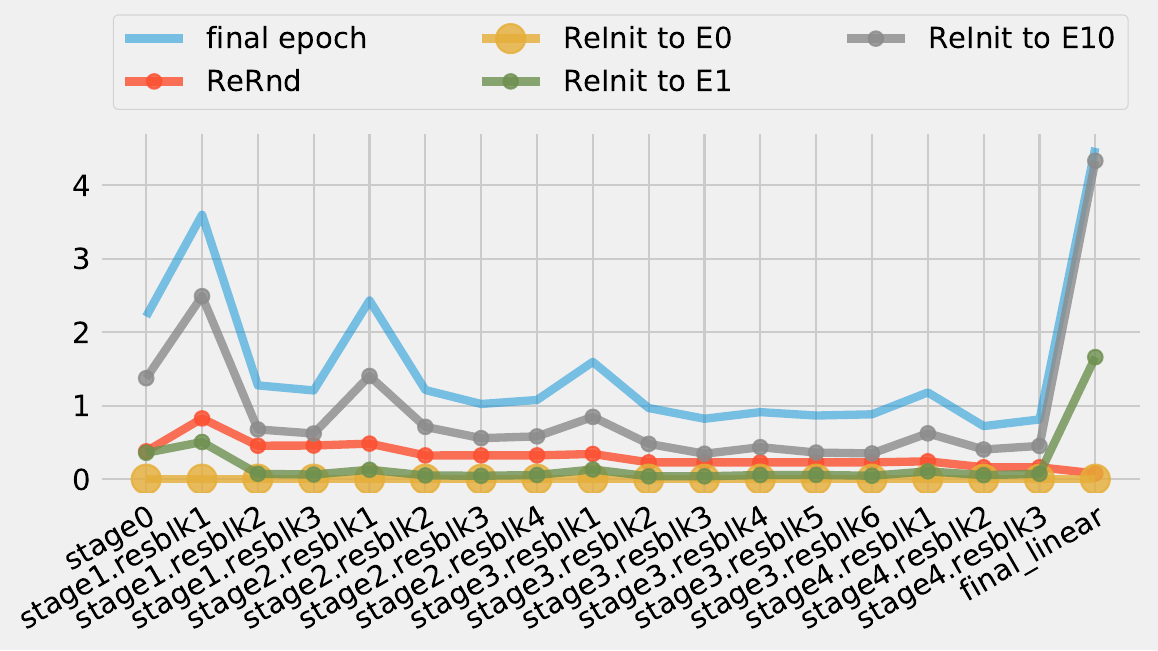}
    \caption{$\|\theta_d^{\tau}-\theta_d^{0}\|_\infty$}
  \end{subfigure}
  \begin{subfigure}[t]{.32\linewidth}
    \includegraphics[width=\linewidth]{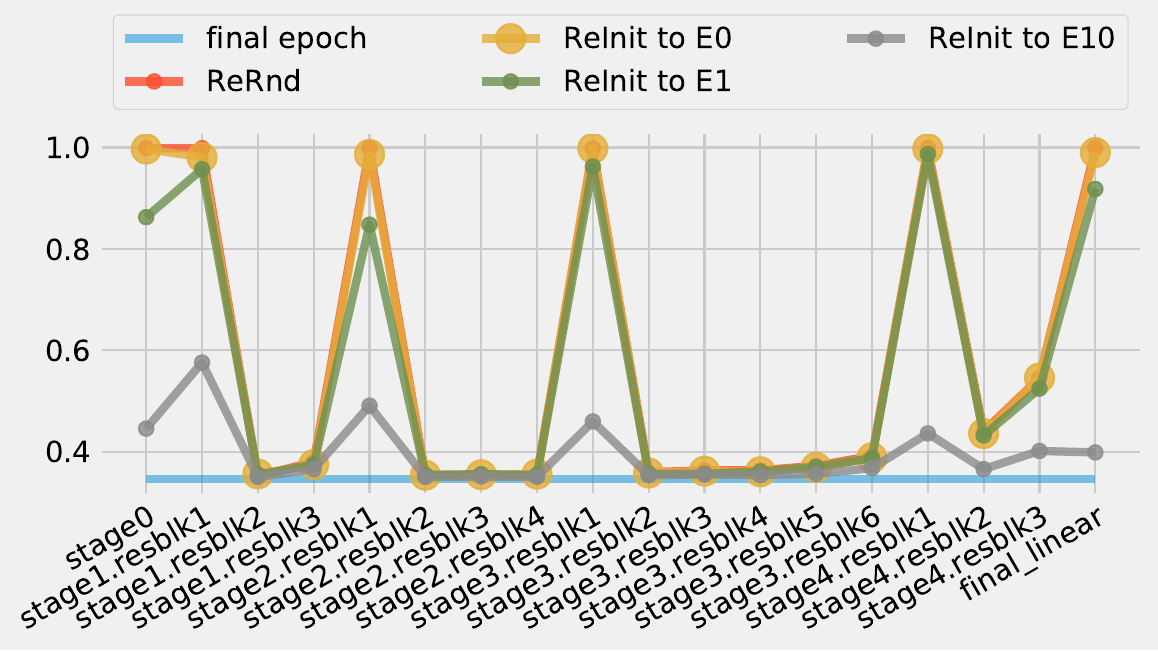}
    \caption{\imagenet test error}
  \end{subfigure}
  \begin{subfigure}[t]{.33\linewidth}
    \includegraphics[width=\linewidth]{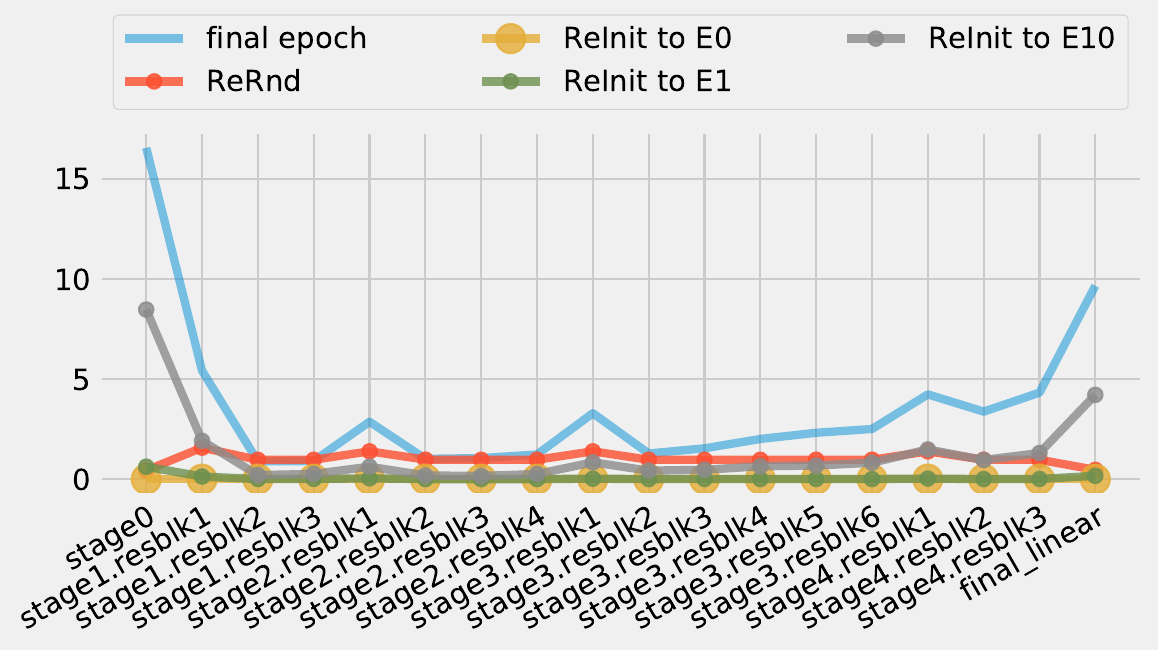}
    \caption{$\|\theta_d^{\tau}-\theta_d^{0}\|$}
  \end{subfigure}
  \begin{subfigure}[t]{.33\linewidth}
    \includegraphics[width=\linewidth]{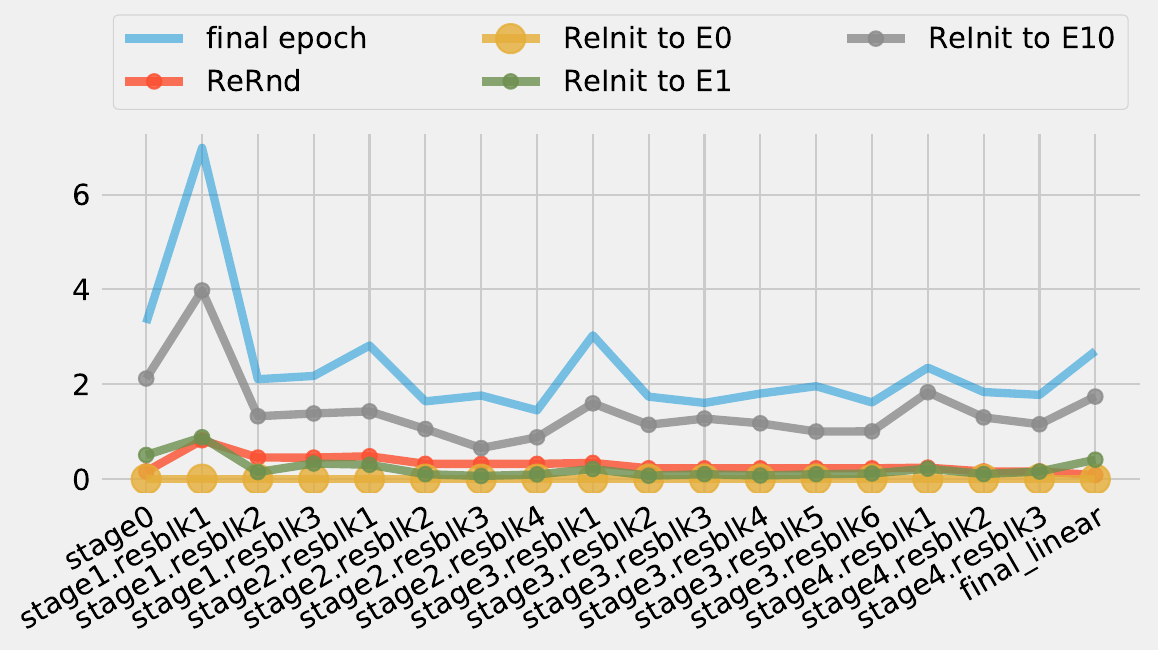}
    \caption{$\|\theta_d^{\tau}-\theta_d^{0}\|_\infty$}
  \end{subfigure}
  \caption{\textbf{Alternative visualization of layer robustness analysis for \resnet{-50} on \cifar (first row) and \imagenet (second row)}.}
  \label{fig:app-reset-curves-res50}
\end{figure}

The empirical results on layer robustness are mainly visualized as heatmaps in the main text. The heatmaps allow uncluttered comparison of the results across layers and training epochs. However, it is not easy to tell the difference between numerical values that are close to each other from the color coding. In this section, we provide alternative visualizations that shows the same results with line plots. In particular, Fig.~\ref{fig:app-reset-curves-vgg16} shows the layer robustness analysis for \vgg{-16} on \cifar{}. Fig.~\ref{fig:app-reset-curves-res50} shows the results for \resnet{-50} on \cifar{} and \imagenet{}, respectively.

\bibliography{refs}

\end{document}